\documentclass[lettersize,journal]{IEEEtran}

\usepackage{amsmath,amsfonts,bm}

\def\eqref#1{equation~\ref{#1}}

\def\1{\bm{1}}

\def\ra{{\textnormal{a}}}

\def\ro{{\textnormal{o}}}

\def\rr{{\textnormal{r}}}

\def\rva{{\mathbf{a}}}

\def\rvo{{\mathbf{o}}}

\def\va{{\bm{a}}}

\def\vo{{\bm{o}}}

\DeclareMathAlphabet{\mathsfit}{\encodingdefault}{\sfdefault}{m}{sl}
\SetMathAlphabet{\mathsfit}{bold}{\encodingdefault}{\sfdefault}{bx}{n}

\def\gK{{\mathcal{K}}}

\DeclareMathOperator*{\argmax}{arg\,max}
\DeclareMathOperator*{\argmin}{arg\,min}

\newcommand{\te}[1]{\texttt{#1}}

\usepackage{amsmath,amsfonts}
\usepackage{algorithmic}
\usepackage{algorithm}
\usepackage{array}
\usepackage{textcomp}
\usepackage{stfloats}
\usepackage{url}
\usepackage{verbatim}
\usepackage{graphicx}
\usepackage{cite}
\usepackage{multirow}
\usepackage{colortbl}
\usepackage{xcolor}
\usepackage{float}
\usepackage[pagebackref,breaklinks,colorlinks]{hyperref}
\hypersetup{citecolor=[RGB]{119,185,0}}
\usepackage[capitalize,noabbrev]{cleveref}
\usepackage[numbers, compress]{natbib}
\usepackage{breakcites}
\usepackage{amssymb}
\usepackage[T1]{fontenc}
\usepackage{duckuments}
\usepackage{tabularx,booktabs}
\usepackage{pifont}
\usepackage{wrapfig}
\usepackage{enumitem}
\usepackage{subcaption}
\usepackage{bbding}

\newcommand{\PreserveBackslash}[1]{\let\temp=\\#1\let\\=\temp}
\newcolumntype{C}[1]{>{\PreserveBackslash\centering}p{#1}}
\newcolumntype{R}[1]{>{\PreserveBackslash\raggedleft}p{#1}}
\newcolumntype{L}[1]{>{\PreserveBackslash\raggedright}p{#1}}

\begin{document}

\title{Communication Learning in Multi-Agent Systems\\ from  Graph Modeling Perspective}

\author{Shengchao~Hu,
        ~Li~Shen, 
        Ya Zhang, 
        ~and~Dacheng~Tao,~\IEEEmembership{Fellow,~IEEE}%
\thanks{Shengchao Hu and Ya Zhang are with Shanghai Jiao Tong University and Shanghai AI Lab, China. Email: \{charles-hu,ya\_zhang\}@sjtu.edu.cn}
\thanks{Li Shen is with Shenzhen Campus of Sun Yat-sen University, Shenzhen 518107, China.
 Email: mathshenli@gmail.com}
\thanks{Dacheng Tao is with Nanyang Technological University, Singapore.
Email: dacheng.tao@ntu.edu.sg}
}

\maketitle

\begin{abstract}
    In numerous artificial intelligence applications, the collaborative efforts of multiple intelligent agents are imperative for the successful attainment of target objectives.
    To enhance coordination among these agents, a distributed communication framework is often employed, wherein each agent must be capable of encoding information received from the environment and determining how to share it with other agents as required by the task at hand.
    However, indiscriminate information sharing among all agents can be resource-intensive, and the adoption of manually pre-defined communication architectures imposes constraints on inter-agent communication, thus limiting the potential for effective collaboration.
    Moreover, the communication framework often remains static during inference, which may result in sustained high resource consumption, as in most cases, only key decisions necessitate information sharing among agents.
    In this study, we introduce a novel approach wherein we conceptualize the communication architecture among agents as a learnable graph. 
    We formulate this problem as the task of determining the communication graph while enabling the architecture parameters to update normally, thus necessitating a bi-level optimization process. 
    Utilizing continuous relaxation of the graph representation and incorporating attention units, our proposed approach, CommFormer, efficiently optimizes the communication graph and concurrently refines architectural parameters through gradient descent in an end-to-end manner.
    Additionally, we introduce a temporal gating mechanism for each agent, enabling dynamic decisions on whether to receive shared information at a given time, based on current observations, thus improving decision-making efficiency.
    Extensive experiments on a variety of cooperative tasks substantiate the robustness of our model across diverse cooperative scenarios, where agents are able to develop more coordinated and sophisticated strategies regardless of changes in the number of agents.
\end{abstract}


\section{Introduction}
\IEEEPARstart{M}{ulti-agent} Reinforcement Learning (MARL) algorithms play an essential role in solving complex decision-making tasks through the analysis of interaction data between computerized agents and simulated or physical environments.
This paradigm finds prevalent application across domains, including autonomous driving \citep{zhou2020smarts, hu2022st}, order dispatching \citep{li2019efficient, yang2018mean}, and gaming AI systems \citep{peng2017multiagent, zhou2023malib}.
In the MARL scenarios typically explored in these studies, multiple agents engage in iterative interactions within a shared environment, continually refining their policies through learning from observations to collectively attain a common objective.
This problem can be conceptually simplified as an instance of independent RL, wherein each agent regards other agents as elements of its environment. 
However, the strategies employed by other agents exhibit dynamic uncertainty and evolve throughout the training process, rendering the environment intrinsically unstable from the viewpoint of each individual agent. 
Consequently, effective collaboration among agents becomes a formidable challenge. 
Additionally, it's important to note that policies acquired through independent RL are susceptible to overfitting with respect to the policies of other agents, as evidenced by \citet{lanctot2017unified}.

Communication is a fundamental pillar in addressing this challenge, serving as a cornerstone of intelligence by enabling agents to operate cohesively as a collective entity rather than disparate individuals.
Its significance becomes especially apparent when tackling complex real-world tasks where individual agents possess limited capabilities and restricted visibility of the environment \citep{lajoie2021towards, yu2022surprising, liu2021coach}.
In this work, we consider MARL scenarios wherein the task at hand is of a cooperative nature and agents are situated in a partially observable environment, but each is endowed with different observation power.
Each agent must not only encode information received from the environment but also determine when and how to share this information with other agents based on the requirements of the task.
Each agent is underpinned by a deep feed-forward network, augmented with access to a communication channel conveying continuous vectors.
Considering bandwidth-related constraints, particularly in instances involving wireless communication channels, a limited subset of agents is permitted to exchange messages during each time step to ensure reliable message transfer \citep{kim2019learning}.
Thus, agents must carefully consider both the content of the messages they transmit and the choice of recipient agents, balancing the need for effective collaboration with the constraints of the communication medium.

\begin{figure*}
    \centering
    \includegraphics[width=0.9\linewidth]{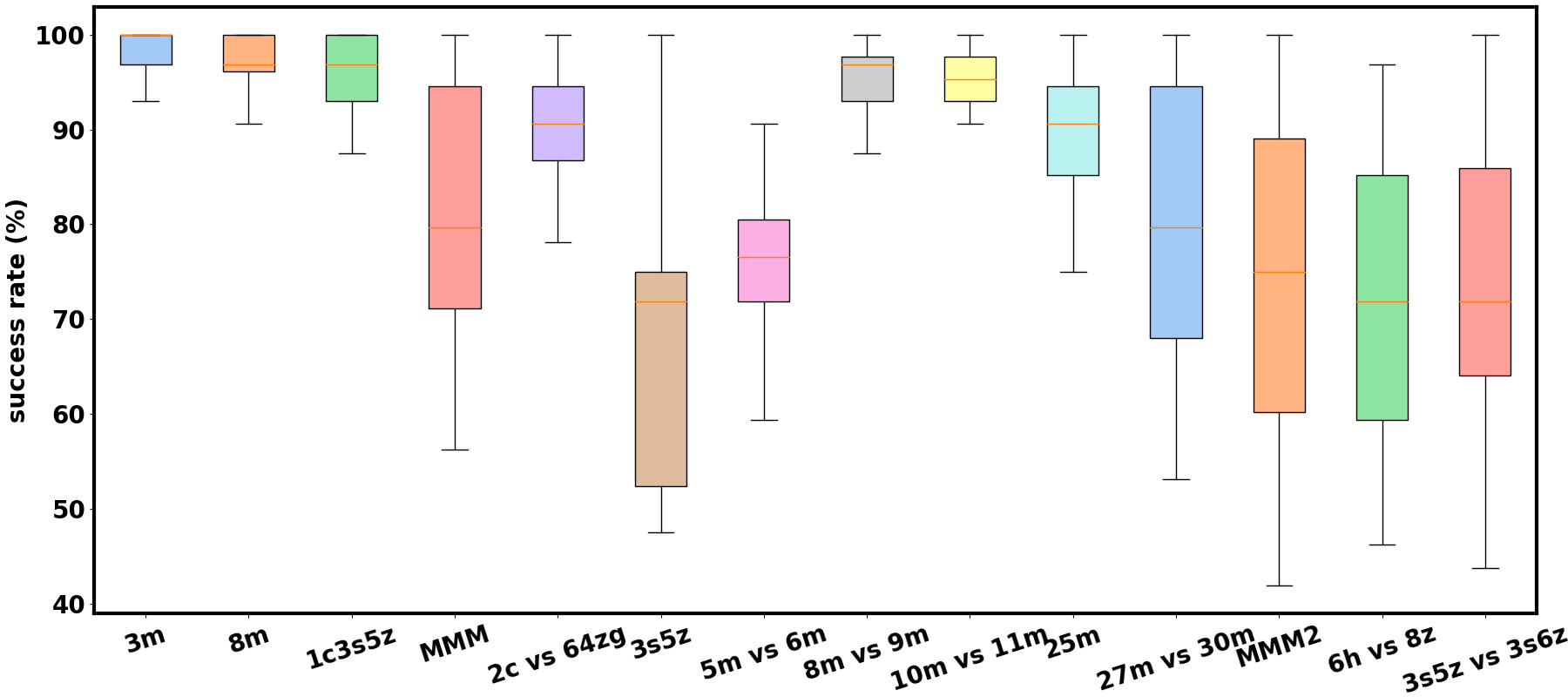}
    \caption{
    The performance of pre-defined communication architectures evaluated across various StarCraft II combat scenarios, with each scenario utilizing ten distinct architectures generated from different random seeds. 
    The significant variance in performance metrics highlights the influence of communication architecture on the agents' effectiveness in these complex environments, emphasizing the necessity of searching for the optimal communication configuration. }
    \label{fig:variance}
\end{figure*}

To facilitate coordinated message exchange, we adopt the centralized training and distributed execution paradigm, as popularized in recent works such as \citet{foerster2018counterfactual, kuba2021trust, mappo}, which allows agents access to global information and knowledge of opponents' actions during the training phase.
There are several approaches for learning communication in MARL including CommNet \citep{sukhbaatar2016learning}, TarMAC \citep{das2019tarmac}, and ToM2C \citep{wang2021tom2c}.
However, methods relying on information sharing among all agents or relying on manually pre-defined communication architectures can be problematic.
When dealing with a large number of agents, distinguishing valuable information for cooperative decision-making from globally shared data becomes problematic.
In such instances, communication may offer limited benefits or even impede cooperative learning, as excessive and irrelevant information can overwhelm agents \citep{jiang2018learning}.
Moreover, in real-world applications, full-scale communication between all agents can be prohibitively expensive, requiring high bandwidth, introducing communication delays, and imposing significant computational complexity.
Manual pre-defined architectures also suffer from high variance, as evident in Figure \ref{fig:variance}, which underscores the necessity for meticulous architectural design to achieve optimal communication, as randomly designed architectures may inadvertently hinder cooperation and result in poor overall performance.
Certain methods, such as those proposed by \citet{zambaldi2018relational, tacchetti2018relational, mao2020neighborhood}, impose constraints whereby each agent communicates only with its neighbors. However, selecting appropriate neighborhoods in complex applications, where agents perform distinct roles, is often difficult.
Additionally, the communication framework in these methods typically remains static during inference, which may result in unnecessary resource consumption, as only key decisions require information sharing among agents in most cases.
Recently, dynamic adjustments to the communication graph during inference have garnered significant attention in recent MARL research \citep{jiang2018learning, kim2019learning, wang2021tom2c}.
However, these methods assume that any agent could potentially communicate with any other, necessitating the establishment of a communication channel between every pair of agents. This results in an implicit fully connected network, which can still lead to significant resource inefficiencies, particularly in wired communication settings, where each agent must still establish physical connections with others.

To address these challenges, we present a novel approach, named CommFormer, designed to facilitate effective and efficient communication among agents in large-scale MARL within partially observable distributed environments.
We conceptualize the communication structure among agents as a learnable graph and formulate this problem as the task of determining the communication graph while enabling the architecture parameters to update normally, thus necessitating a bi-level optimization process.
In contrast to conventional methods that involve searching through a discrete set of candidate communication architectures, we relax the search space into a continuous domain, enabling architecture optimization via gradient descent in an end-to-end manner.
Diverging from previous approaches that often employ arithmetic or weighted means of internal states before message transmission \citep{peng2017multiagent, wang2021tom2c}, which may compromise communication effectiveness, our method directly transmits each agent's local observations and actions to specific agents based on the learned communication architecture.
Subsequently, each agent employs an attention unit to dynamically allocate credit to received messages from the graph modeling perspective, which enjoys a monotonic performance improvement guarantee \citep{wen2022multi}.
We also introduce a temporal gating mechanism for each agent, enabling dynamic decisions on whether to receive shared information from the sender connected through the communication graph, based on current observations. This approach enhances decision-making efficiency without requiring the establishment of new communication channels.
Extensive experiments conducted in a variety of cooperative tasks substantiate the robustness of our model across diverse cooperative scenarios. 
CommFormer with static inference consistently outperforms strong baselines and achieves comparable performance to methods that permit unrestricted information sharing among all agents, demonstrating its effectiveness regardless of variations in the number of agents.
When implemented with the temporal gating mechanism, CommFormer still maintains performance levels akin to those of static methods, thereby highlighting the efficiency of our dynamic approach, which substantially conserves communication resources.

Our contributions can be summarized as follows:

\begin{itemize}[leftmargin=*]
    \item We conceptualize the communication structure as a graph and introduce an innovative algorithm for learning it through bi-level optimization, which efficiently enables the simultaneous optimization of the communication graph and architectural parameters.
    \item We propose the adoption of the attention unit within the framework of graph modeling to dynamically allocate credit to received messages, thereby enjoying a monotonic performance improvement guarantee while also improving communication efficiency. 
    \item We introduce a temporal gating mechanism for each agent, allowing for dynamic decisions regarding the reception of shared information based on current observations, which enhances decision-making efficiency without necessitating the establishment of new communication channels.
    \item Through extensive experiments on a variety of cooperative tasks, CommFormer consistently outperforms robust baseline methods and achieves performance levels comparable to approaches that permit unrestricted information sharing among all agents.
\end{itemize}

\section{Related Work}

\textbf{Multi-agent Cooperation}.
As a natural extension of single-agent RL, MARL has garnered considerable attention for addressing complex problems within the framework of Markov Games \citep{yang2020overview}.
Numerous MARL methodologies have been developed to tackle cooperative tasks in an online setting, where all participating agents collaborate toward a shared reward objective.
While this cooperative paradigm is crucial, it also introduces significant challenges in distributed systems. 
Throughout the training process, agents' policies constantly evolve, leading to a non-stationary environment and posing difficulties for model convergence.
To address the challenge of non-stationarity in MARL, algorithms typically operate within two overarching frameworks: centralized and decentralized learning.
Centralized methods \citep{claus1998dynamics} involve the direct learning of a single policy responsible for generating joint actions for all agents.
On the other hand, decentralized learning \citep{littman1994markov} entails each agent independently optimizing its own reward function. 
While these methods can handle general-sum games, they may encounter instability issues, even in relatively simple matrix games \citep{foerster2017learning}.
Centralized training and decentralized execution (CTDE) algorithms represent a middle ground between these two frameworks. 
One category of CTDE algorithms is value-decomposition (VD) methods, wherein the joint Q-function is formulated as a function dependent on the individual agents' local Q-functions \citep{rashid2020monotonic, son2019qtran, sunehag2017value}.
The others \citep{lowe2017multi, foerster2018counterfactual} employ actor-critic architectures and learn a centralized critic that takes global information into account.
Researchers have also leveraged the expressive power of Transformers to enhance coordination capabilities, such as MAT \citep{wen2022multi}. 
These methods have demonstrated remarkable cooperation abilities across various tasks, including SMAC \citep{samvelyan2019starcraft}, Hanabi, and GRF \citep{yu2022surprising}.
In this work, we introduce an innovative approach operating under the CTDE paradigm, with limited communication capabilities, which achieves comparable or even superior performance when compared to these established baselines.

\textbf{Communication Learning}.
Communication plays a crucial role in MARL, enabling agents to share experiences, intentions, observations, and other vital information. Three key factors characterize multi-agent communication: when to communicate, which agents to engage with, and the type of information exchanged.
DIAL \citep{foerster2016learning} is the pioneer in proposing learnable communication through back-propagation with deep Q-networks. 
In this method, each agent generates a message at each timestep, which serves as input for other agents in the subsequent timestep. 
Building on this foundational work, a variety of approaches have emerged in the field of multi-agent communication. 
Some methods adopt pre-defined communication architectures, e.g. CommNet \citep{sukhbaatar2016learning}, BiCNet \citep{peng2017multiagent}, and GA-Comm \citep{liu2020multi}. 
These techniques establish fixed communication structures to facilitate information exchange among agents, often utilizing GNN models \citep{niu2021multi, bettini2023heterogeneous}.
In contrast, other approaches such as ATOC \citep{jiang2018learning}, TarMAC \citep{das2019tarmac}, and ToM2C \citep{wang2021tom2c} explore dynamic adaptation of communication structures during inference, enabling agents to selectively transmit or receive information.
Moreover, IC3Net \citep{singh2018learning}, Gated-ACML \citep{mao2020learning}, and I2C \citep{ding2020learning} implement gating mechanisms to determine when and with which teammates to communicate, thereby reducing information redundancy. 
SMS \citep{xue2022efficient} models multi-agent communication as a cooperative game, assessing the significance of each message for decision-making and eliminating channels that do not contribute positively. 
Additionally, CDC \citep{pesce2023learning} dynamically alters the communication graph from a diffusion process perspective, while TWG-Q \citep{liu2022temporal} emphasizes temporal weight learning through the application of weighted graph convolutional networks.
In our research, we establish a fixed communication architecture prior to inference to determine which agents will communicate \citep{CommFormer}, complemented by a dynamic gating mechanism that decides when communication occurs.
Our approach, CommFormer, builds on this concept by learning an optimal communication architecture through back-propagation before the inference phase, thereby determining which teammates to communicate with.
Each agent transmits its local observations and actions as messages through a shared channel, utilizing an attention mechanism to process the received information, thereby defining the type of information exchanged.
Our method also learns a gating network to manage communication efficiently during inference; each agent first decides whether to communicate based on the current observation using the learned gating network, thus determining the timing of communication.

\begin{figure*}
    \centering
    \includegraphics[width=1.0\linewidth]{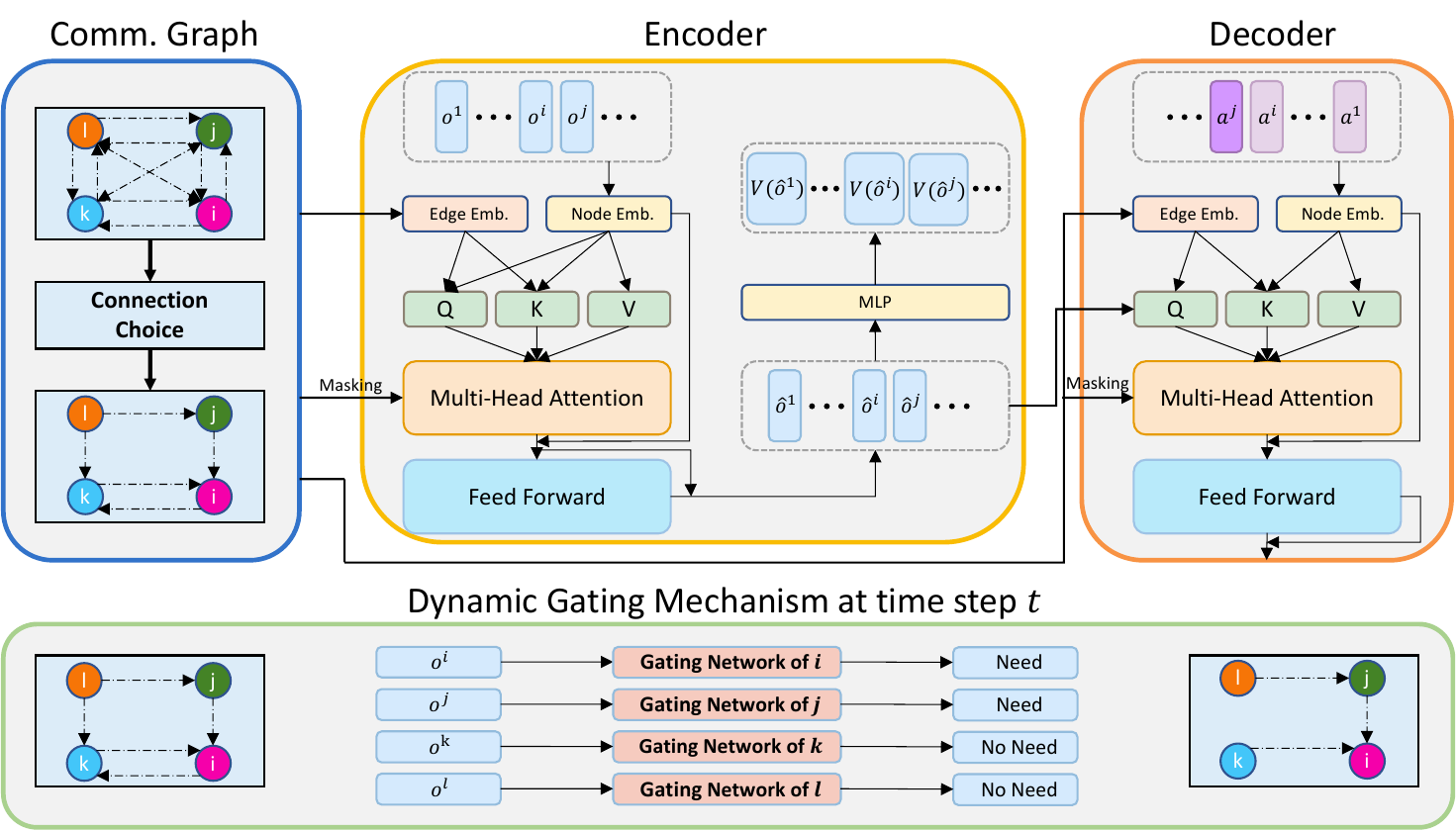}
    \caption{
    The overview of our proposed CommFormer. 
    CommFormer initiates by establishing the communication graph, which subsequently serves as both the masking and edge embeddings in the encoder and decoder to ensure that agents can exclusively access messages from communicated agents.
    Subsequently, the encoder and decoder modules come into play, processing a sequence of agents' observations and transforming them into a sequence of optimal actions.
    Additionally, CommFormer integrates a dynamic gating mechanism that determines when communication is necessary based on the current observations for each agent. 
    For instance, at time step $t$, agents $i$ and $j$ require additional information, while agents $k$ and $l$ do not, resulting in the omission of corresponding edges to conserve resources.
    }
    \label{fig:overview}
\end{figure*}

\section{Method}

The goal of our proposed method is to address the multi-agent collaborative communication problem, which enables agents to operate cohesively as a collective entity rather than disparate individuals.
In this paper, we are interested in \textit{learning to construct the communication graph}, \textit{learning to determine when to communicate}, and \textit{learning how to cooperate with received messages} in a bandwidth-limited way.

\subsection{Problem Formulation}
The MARL problems can be modeled by Markov games $\langle \mathcal{N}, \bm{\mathcal{O}}, \bm{\mathcal{A}}, R, P, \gamma \rangle$ \citep{littman1994markov}.
The set of agents is denoted as $\mathcal{N} = \{1, \dots, n \}$. 
The product of the local observation spaces of the agents forms the joint observation space, denoted as $\bm{\mathcal{O}}=\prod_{i=1}^{n}\mathcal{O}^i$. 
Similarly, the product of the agents' action spaces constitutes the joint action space, represented as $\bm{\mathcal{A}}=\prod_{i=1}^{n}\mathcal{A}^i$. 
The joint reward function, $R:\bm{\mathcal{O}}\times \bm{\mathcal{A}} \rightarrow [-R_{\max}, R_{\max}]$, maps the joint observation and action spaces to the reward range $[-R_{\max}, R_{\max}]$. The transition probability function, $P:\bm{\mathcal{O}}\times \bm{\mathcal{A}} \times \bm{\mathcal{O}} \rightarrow \mathbb{R}$, defines the probability distribution of transitioning from one joint observation and action to another. 
Lastly, the discount factor, denoted as $\gamma \in [0, 1)$, plays a crucial role in discounting future rewards.

At time step $t\in\mathbb{N}$, an agent $i\in\mathcal{N}$ receives an observation denoted as $\ro_{t}^i\in\mathcal{O}^i$. The collection of these individual observations $\vo=(o^1, \dots, o^n)$ forms the "joint" observation. Agent $i$ then selects an action $\ra^{i}_t$ based on its policy $\pi^i$. It's worth noting that $\pi^i$ represents the policy of the $i^{\text{th}}$ agent, which is a component of the agents' joint policy denoted as $\pi$.
Apart from its own local observation $\ro_{t}^i$, each agent possesses the capability to receive observations $\ro_t^j$ from other agents, along with their actions (auto-regressively) $\ra^{j}_t$ through a communication channel. 
At the end of each time step, the entire team collectively receives a joint reward denoted as $R(\rvo_{t}, \rva_t)$ and observes $\rvo_{t+1}$, following a probability distribution $P(\cdot|\rvo_{t}, \rva_t)$.
Over an infinite sequence of such steps, the agents accumulate a discounted cumulative return denoted as $R^{\gamma} \triangleq \sum_{t=0}^{\infty}\gamma^t R(\rvo_{t}, \rva_t)$.

In practical scenarios where agents have the capability to communicate with each other over a shared medium, two critical constraints are imposed: bandwidth and contention for medium access \citep{kim2019learning}.
The bandwidth constraint implies that there is a limited capacity for transmitting bits per unit time, and the contention constraint necessitates the avoidance of collisions among multiple transmissions, which is a natural aspect of signal broadcasting in wireless communication. 
Consequently, each agent can only transmit their message to a restricted number of other agents during each time step to ensure reliable message transfer.
In this paper, we conceptualize the communication architecture as a directed graph, denoted as $\mathcal{G} = \langle \mathcal{V}, \mathcal{E} \rangle$, where each node $v_i \in \mathcal{V}$ represents an agent, and an edge $e_{i \rightarrow j} \in \mathcal{E}$ signifies message passing from agent $v_i$ to agent $v_j$.
The restriction on communication can be mathematically expressed as the sparsity $\mathcal{S}$ of the adjacency matrix of the edge connections $\alpha$. 
This sparsity parameter, $\mathcal{S}$, controls the allowed number of connected edges, which is given by $\mathcal{S} \times N^2$, where $N$ is the number of agents.

\subsection{Architecture}
The overall architecture of our proposed CommFormer is illustrated in Figure \ref{fig:overview}.

\textbf{Communication Graph.} To design a communication-efficient MARL paradigm, we introduce the Communication Transformer or CommFormer, which adopts a graph modeling paradigm, inspired by developments in sequence modeling \citep{hu2022transforming, hu2023graph, HarmoDT, QT, wen2022multi}.
We apply the Transformer architecture which facilitates the mapping between the input, consisting of agents' observation sequences $(o^{1}, \dots, o^{n})$, and the output, which comprises agents' action sequences $(a^{1}, \dots, a^{n})$.
Considering communication constraints, each agent has a limited capacity to communicate with a subset of other agents, represented by the sparsity $\mathcal{S}$ of the adjacency matrix of the edge connections.
To identify the optimal communication graph, we treat multiple agents as nodes in a graph and introduce a learnable adjacency matrix, represented by the parameter matrix $\alpha \in \mathbb{R}^{N \times N}$, which are optimized through gradient descent during training in an end-to-end manner.

\textbf{Encoder.} 
The encoder, whose parameters are denoted by $\phi$, takes a sequence of observations $(o^{1}, \dots, o^{n})$ as input and passes them through several computational blocks.
Each such block consists of a \textsl{relation-enhanced} mechanism \citep{hu2023graph, cai2020graph} and a \textsl{multi-layer perceptron} (MLP), as well as \textsl{residual connections} to prevent gradient vanishing and network degradation with the increase of depth. 
In the vanilla multi-head attention, the attention score between the element $o^i$ and $o^j$ can be formulated as the dot-product between their query vector and key vector:
\begin{equation}
    s_{ij} = f(o^i, o^j) = o^i W_q^T W_k o^j.
\end{equation}
$s_{ij}$ can be regarded as implicit information associated with the edge $e_{j \rightarrow i}$, where agent $i$ queries the information sent from agent $j$.
To identify the most influential edge contributing to the final performance, we augment the implicit attention score with explicit edge information:
\begin{equation}
\label{eq:attn_score}
    \begin{split}
        s_{ij} &= g(o^i, o^j, r_{i \rightarrow j}, r_{j \rightarrow i}) \\
            &= (o^i + r_{i \rightarrow j})W_q^TW_k (o^j + r_{j \rightarrow i} ),
    \end{split}
\end{equation}
where $r_{* \rightarrow *}$ is obtained from an embedding layer that takes the adjacency matrix $\alpha$ as input.
We also apply a mask to the attention scores using the adjacency matrix $\alpha$ to ensure that only information from connected agents is accessible:
\begin{equation}
\label{eq:mask}
    s_{ij} = 
    \begin{cases}
        s_{ij}, & e_{j\rightarrow i} = 1, \\
        -\infty, & e_{j\rightarrow i} = 0. 
    \end{cases}
\end{equation}
We represent the encoded observations as $(\hat{\vo}^{1}, \dots, \hat{\vo}^{n})$, which capture not only the individual agent information but also the higher-level inter-dependencies between agents through communication.
To facilitate the learning of expressive representations, during the training phase, we treat the encoder as the critic and introduce an additional projection to estimate the value functions:
\begin{equation}
\label{eq:encoder-loss}
\begin{split}
    L_{\text{Encoder}}(\phi) = \frac{1}{Tn} &\sum_{m=1}^{n}\sum_{t=0}^{T-1}\Big[ R(\rvo_t, \rva_t) + \\
    & \gamma V_{\bar{\phi}}(\hat{\rvo}^{{m}}_{t+1}) - V_{\phi}(\hat{\rvo}^{{m}}_t)\Big]^2,
\end{split}
\end{equation}
where $\bar{\phi}$ is the target network's parameter, which is a separate neural network that is a copy of the main value function. 
The update mechanism for $\bar{\phi}$ is executed either through an exponential moving average or via periodic updates in a "hard" manner, typically occurring every few epochs \citep{mnih2015human}.

\textbf{Decoder.} 
The decoder, characterized by its parameters $\theta$, processes the embedded joint action $\va^{{0:m-1}}, m={1, \dots n}$ through a series of decoding blocks.
The decoding block also incorporates a \textsl{relation-enhanced} mechanism for calculating attention between encoded actions and observation representations, along with an MLP and \textsl{residual connections}.
In addition to the adjacency matrix mask, we apply a constraint that limits attention computation to occur only between agent $i$ and its preceding agents $j$ where $j < i$. 
This constraint maintains the sequential update scheme, ensuring that the decoder produces the action sequence in an auto-regressive manner: $\pi^{m}_{\theta}(\ra^{m}|\hat{\rvo}^{{1:n}}, \rva^{{1:m-1}})$, which guarantees monotonic performance improvement during training \citep{wen2022multi}.
We apply the PPO algorithm \citep{schulman2017proximal} to train the decoder agent:
\begin{equation}
\label{eq:decoder-loss}
    \begin{split}
        L_{\text{Decoder}}(\theta) &= -\frac{1}{Tn}\sum_{m=1}^{n}\sum_{t=0}^{T-1} \\
        \min &\Big( \rr^{m}_{t}(\theta)\hat{A}_t, \text{clip}(\rr^{m}_{t}(\theta), 1\pm \epsilon)\hat{A}_t \Big), \\
        \rr^{m}_{t}(\theta) &= \frac{\pi^{m}_{\theta}(\ra^{m}_t|\hat{\rvo}^{1:n}_t, \hat{\rva}^{1:m-1}_t)}{\pi^{m}_{\theta_{\text{old}}}(\ra^{m}_t|\hat{\rvo}^{1:n}_t, \hat{\rva}^{1:m-1}_t)}, 
    \end{split}
\end{equation}
where $\hat{A}_t$ is an estimate of the joint advantage function, which can be formulated as $\hat{V}_t = \frac{1}{n}\sum_{m=1}^{n}V(\hat{\ro}^{m}_t)$ using the generalized advantage estimation (GAE) approach \citep{schulman2015high}.

\textbf{Dynamic Gating.} 
In many situations, only critical decisions necessitate information sharing among agents, while most interactions do not require additional information. 
To address this, we develop a dynamic gating mechanism for each agent to determine whether to receive shared information from the sender, as specified by the communication graph, based on current observations.
Specifically, each agent is equipped with a gating network that shares the same architecture but maintains distinct parameters.
This gating network is constructed using two linear layers, along with one GeLU activation function and one Layer Normalization:
\begin{equation}
\label{eq:dyn_h}
    h^i_t = \Psi^{\text{gate}}(o^i_t | \kappa_i) ~~ \text{for} ~~ i=1,\dots, n,
\end{equation}
where $o^i_t$ is the local observation for agent $i$ at time step $t$, $\kappa_i$ represents the parameters for the gating network of agent $i$, and $\mathcal{K} = [\kappa_1, \dots, \kappa_n]$ denotes the set of parameters for all agents' gating networks. 
At time step $t$, the communication graph is then updated as follows:
\begin{equation}
\label{eq:dyn_alpha}
    \alpha^{\text{dyn}}_i = \alpha_i \times h^i_t ~~ \text{for} ~~ i=1,\dots, n,
\end{equation}
where $\alpha_i$ is the $i$-th row of the communication matrix $\alpha$.

\subsection{Training and Execution}

We employ the CTDE paradigm: during centralized training, there are no restrictions on communication between agents.
However, once the learned policies are executed in a decentralized manner, agents can only communicate through a constrained bandwidth channel.
We employ a two-stage training pipeline. In the first stage, we focus on learning the communication graph. Once this graph is established, the second stage involves training the dynamic gating mechanism to determine optimal communication timing based on the learned communication graph.

\begin{algorithm*}[!htbp]
    \caption{CommFormer}
    \begin{algorithmic}[1]
    
    \STATE \textbf{Input:} Batch size $B$, number of agents $N$, episodes $K$, steps per episode $T$, sparsity $\mathcal{S}$.\\
    \STATE \textbf{Initialize:} Encoder $\{\phi\}$, Decoder $\{\theta\}$, Dynamic Gating $\{ \gK \}$, Replay buffer $\mathcal{B}$, Adjacency matrix $\alpha \in \mathbb{R}^{n \times n}$. \\
    
    \FOR{$k = 0,1, \dots, K-1$}
        \FOR{$t = 0,1, \dots, T-1$}
            \STATE Collect a sequence of observations $o^{{1}}_t, \dots, o^{{n}}_t$ from environments.\\
            \STATE {\color{gray}\te{// inference with CommFormer}} \\
            \STATE Generate the matrix $e \in \{ 0, 1\}^{n \times n}$ according to the $\alpha$ with Equation \ref{eq:sample_exact}.
            \STATE {\color{gray}\te{// If with dynamic inference}}
            \STATE Update the matrix $e$ based on the current observations with the dynamic gating networks by Equations \ref{eq:dyn_h} and \ref{eq:dyn_alpha}.
            \STATE Generate representation sequence $\hat{o}^{{1}}_t, \dots, \hat{o}^{{n}}_t$ via Encoder $\phi$ with attention score (Equation \ref{eq:attn_score}) and mask (Equation \ref{eq:mask}), similar to the Decoder. \\
            \FOR{$m = 0, 1, \dots, n-1$}
                \STATE Input $\hat{o}^{{1}}_t, \dots, \hat{o}^{{n}}_t$ and $a^{0}_t, \dots, a^{m}_t$ to the Decoder $\theta$ and infer $a^{{m+1}}_t$ with the auto-regressive manner. \\
            \ENDFOR
            \STATE Execute joint actions $a^{0}_t, \dots, a^{n}_t$ in environments and collect the reward $R(\vo_t, \va_t)$.\\
            \STATE Insert $(\vo_{t}, \va_{t},R (\vo_t, \va_t))$ in to $\mathcal{B}$.\\
        \ENDFOR\\
        \STATE {\color{gray}\te{// train the CommFormer}} \\
        \STATE Sample a random minibatch of $B$ steps from $\mathcal{B}$.
        \STATE Generate the matrix $e \in \{ 0, 1\}^{n \times n}$ according to the $\alpha$ with Equation \ref{eq:sample_gumbel}.
        \STATE Generate $V_\phi(\hat{o})$ with the output layer of the Encoder $\phi$ and compute the joint advantage function $\hat{A}$ based on $V_\phi(\hat{o})$ with GAE.\\
        \STATE Input $\hat{o}^{{1}}, \dots, \hat{o}^{{n}}$ and $a^{0}, \dots, a^{{n-1}}$, generate $\pi^{1}_{\theta}, \dots, \pi^{n}_{\theta}$ at once with the Decoder $\theta$.\\
        \STATE Calculate the training loss $L = L_{\text{Encoder}}(\phi) + L_{\text{Decoder}}(\theta)$ with Equation 
        \ref{eq:encoder-loss} and Equation \ref{eq:decoder-loss}.\\
        \STATE Iteratively update the $\phi, \theta$ and $\alpha$ with Equation \ref{eq:weight_loss} and Equation \ref{eq:structure_loss}.
        \STATE {\color{gray}\te{// train the dynamic gating networks}} \\
        \STATE Generate the matrix $e \in \{ 0, 1\}^{n \times n}$ according to the $\alpha$ with Equation \ref{eq:sample_exact}.
        \STATE Update the matrix $e$ based on the current observations with the dynamic gating networks by Equations \ref{eq:dyn_h} and \ref{eq:dyn_alpha}.
        \STATE Repeat steps 20-23, replacing $\alpha$ with $\gK$ for the subsequent iterations.
    \ENDFOR
    \end{algorithmic}
    \label{alg:commformer}
\end{algorithm*}

\subsubsection{Centralized Training}

During the first stage of the training process, we need to determine the communication matrix $\alpha$ while allowing the architecture parameters $\phi$ and $\theta$ to update normally.
This implies a bi-level optimization problem \citep{anandalingam1992hierarchical, colson2007overview} with $\alpha$ as the upper level variable and $\phi$ and $\theta$ as the lower-level variable:
\begin{align}
    \min_{\alpha}~~~ & \mathcal{L}_{val} (\phi^*(\alpha), \theta^*(\alpha), \alpha), \\
    \text{s.t.}~~~ & \phi^*(\alpha), \theta^*(\alpha) = \argmin_{\phi, \theta} \mathcal{L}_{train}(\phi, \theta, \alpha) \label{eq_inner}, \\
    & |\alpha| \leq \mathcal{S} \times N^2 \label{eq_sparsity},
\end{align}
where $\mathcal{L} = L_{\text{Encoder}}(\phi) + L_{\text{Decoder}}(\theta)$ with different online rollouts for training $L_{train}$ and validation $L_{val}$, and $|\alpha|$ denotes the number of connected edges.
Evaluating the architecture gradient exactly can be prohibitive due to the expensive inner optimization, and each value in $\alpha$ is represented by a discrete value in $\{0, 1\}$.
We propose a simple approximation scheme that alternately updates the following formula and relaxes $\alpha$ as a continuous matrix to enable differentiable updating:
\begin{equation}
\label{eq:weight_loss}
\begin{split}
    \phi &= \phi - \xi \nabla_{\phi} \mathcal{L}_{train}(\phi, \theta, \alpha) \\
    \theta &= \theta - \xi \nabla_{\theta} \mathcal{L}_{train} (\phi, \theta, \alpha),
\end{split}
\end{equation}
and 
\begin{equation}
\label{eq:structure_loss}
    \alpha = \alpha - \eta \nabla_{\alpha} \mathcal{L}_{val} (\phi, \theta, \alpha),
\end{equation}
where $\phi, \theta$ denote the current weights maintained by the algorithm, and $\xi, \eta$ are the learning rate for a step of inner and outer optimization.
The idea is to approximate $\phi^*(\alpha), \theta^*(\alpha)$ by adapting $\phi$ and $\theta$ using only a single training step, without fully solving the inner optimization (Equation \ref{eq_inner}) by training until convergence.

To update the discrete adjacency matrix $\alpha$, we utilize the Gumbel-Max trick \citep{jang2016categorical, maddison2016concrete} to sample the binary adjacency matrix, which facilitates the continuous representation of $\alpha$ and enables the normal back-propagation of gradients during training.
To satisfy constraint \ref{eq_sparsity}, we extend the original one-hot Gumbel-Max trick to k-hot, enabling each agent to send messages to a fixed number of $k = \mathcal{S} \times N$ agents:
\begin{equation}
\label{eq:sample_gumbel}
\begin{split}
    e_i = \text{k\_hot} &\big( \text{k-}\argmax [\text{Softmax}(\alpha_{ij} + g_j), \\
    & \text{for} ~ j=1,\dots,n ] \big), 
\end{split}
\end{equation}
where $g_j$ is sampled from Gumbel(0,1), and $e_i \in \mathbb{N}^N$ represents the edges connected to agent $i$.

During the second stage of the training process, once the communication graph has been effectively learned, we focus on optimizing the dynamic gating networks with a similar objective function:
\begin{align}
    \min_{\gK}~~~ & \mathcal{L}_{val} (\phi^*(\gK), \theta^*(\gK), \gK), \\
    \text{s.t.}~~~ & \phi^*(\gK), \theta^*(\gK) = \argmin_{\phi, \theta} \mathcal{L}_{train}(\phi, \theta, \gK) \label{eq_inner2}.
\end{align}
In this formulation, the parameters of the communication matrix $\alpha$ remain fixed, while $\gK$ influences performance by dynamically altering the communication graph as defined by Equations \ref{eq:dyn_h} and \ref{eq:dyn_alpha}. 
This allows us to employ a similar bi-level optimization strategy, utilizing the Gumbel-Max trick to simultaneously update the parameters of the dynamic gating networks $\gK$ and the architecture parameters $\phi$ and $\theta$.

\subsubsection{Distributed Execution}

During execution, each agent $i$ has access to its local observations and actions, as well as additional information transmitted by other agents through communication.
The adjacency matrix is derived from the parameters $\alpha$ without any randomness as follows:
\begin{equation}
\label{eq:sample_exact}
    e_i = \text{k\_hot} \big( \text{k-}\argmax \left[\alpha_{ij}, ~ \text{for} ~ j=1,\dots,n \right] \big).
\end{equation}
In the case of dynamic inference, the adjacency matrix undergoes an additional masking process at time step $t$, defined by the following equation:
\begin{equation}
    e_i^{\text{dyn}} = e_i \times h_t^i,
\end{equation}
where $h_t^i$ is the dynamic gating mechanism's output at time step $t$, as specified in Equation \ref{eq:dyn_h}.
Note that each action is generated auto-regressively, in the sense that $a^m$ will be inserted back into the decoder again to generate $a^{m+1}$ (starting with $a^1$ and ending with $a^{n}$).
Through the use of limited communication, each agent is still able to effectively select actions when compared to fully connected agents, which leads to significant reductions in communication costs and overhead.
The overall pseudocode is presented in Algorithm \ref{alg:commformer}.

\section{Experiment}

To evaluate the properties and performance of our proposed CommFormer\footnote{Our code is available at: \url{https://github.com/charleshsc/CommFormer}}, we conduct a series of experiments using four environments, including Predator-Prey (PP) \citep{singh2018learning}, Predator-Capture-Prey (PCP) \citep{seraj2022learning}, StarCraftII Multi-Agent Challenge (SMAC) \citep{samvelyan2019starcraft}, and Google Research Football(GRF) \citep{kurach2020google}.
\begin{itemize}[leftmargin=*]
    \item \textbf{PP}.  
    The goal is for $N$ predator agents with limited vision to find a stationary prey and move to its location. The agents in this domain all belong to the same class (i.e., identical state, observation and action spaces).

    \item \textbf{PCP}.
     We have two classes of predator and capture agents. Agents of the predator class have the goal of finding the prey with limited vision (similar to agents in PP). Agents of the capture class, have the goal of locating the prey and capturing it with an additional capture-prey action in their action-space, while not having any observation inputs (e.g., lack of scanning sensors).

    \item \textbf{SMAC}. 
    In these experiments, CommFormer controls a group of agents tasked with defeating enemy units controlled by the built-in AI. 
    The level of combat difficulty can be adjusted by varying the unit types and the number of units on both sides. 
    We measure the winning rate and compare it with state-of-the-art baseline approaches.
    Notably, the maps \textit{1o10b\_vs\_1r} and \textit{1o2r\_vs\_4r} present formidable challenges attributed to limited observational scope, while the map \textit{5z\_vs\_1ul} necessitates heightened levels of coordination to attain successful outcomes.

    \item \textbf{GRF}.
    We evaluate algorithms in the football \textit{academy scenario 3 vs. 2}, where we have 3 attackers vs. 1 defender, and 1 goalie. The three offending agents are controlled by the MARL algorithm, and the two defending agents are controlled by a built-in AI. We find that utilizing a 3 vs. 2 scenario challenges the robustness of MARL algorithms to stochasticity and sparse rewards.
\end{itemize}
It is worth noting that in certain domains, our objective extends beyond maximizing the average success rate or cumulative rewards. We also aim to minimize the average number of steps required to complete an episode, emphasizing the ability to achieve goals in the shortest possible time.

\begin{table*}[t!]
\caption{
Performance evaluations of different metrics and standard deviation on the selected benchmark, where UPDeT's official codebase supports several Marine-based tasks only.
In this context, CF refers to CommFormer, while CF-dyn denotes CommFormer-dyn. 
Note that the sparsity parameter $\mathcal{S}$ in CommFormer is consistently set to 0.4 for all tasks evaluated. 
}
 \renewcommand{\arraystretch}{1.0}
  \centering
    \resizebox{0.97\textwidth}{!}{
    \begin{tabular}{cc|cccccccc|c}
    \toprule
    Task & Difficulty & CF-dyn(0.4) & CF(0.4) &  MAT & MAPPO & HAPPO & QMIX & UPDeT &FC & Steps\\
    \midrule
    \tiny{3m} & Easy & \textbf{100.0}\tiny{(0.4)} & \textbf{100.0}\tiny{(0.0)} &  \textbf{100.0}\tiny{(0.0)} & \textbf{100.0}\tiny{(0.4)} & \textbf{100.0}\tiny{(1.2)} & 96.9\tiny{1.3} & \textbf{100.0}\tiny{(5.2)} & \textbf{100.0}\tiny{(0.0)} & 5e5\\
    \tiny{8m} & Easy & \textbf{100.0}\tiny{(0.0)} & \textbf{100.0}\tiny{(0.0)} & \textbf{100.0}\tiny{(0.0)} & 96.8\tiny{(2.9)} & 97.5\tiny{(1.1)} & 97.7\tiny{1.9} & 96.3\tiny{(9.7)}& \textbf{100.0}\tiny{(0.0)} & 1e6\\
    \tiny{1c3s5z} & Easy & \textbf{100.0}\tiny{(0.0)} & \textbf{100.0}\tiny{(0.0)} & \textbf{100.0}\tiny{(0.0)} & \textbf{100.0}\tiny{(2.2)} & 97.5\tiny{(1.8)} & 96.9\tiny{(1.5)} & / & \textbf{100.0}\tiny{(0.0)} & 2e6\\
    \tiny{MMM} & Easy & \textbf{100.0}\tiny{(0.0)} & \textbf{100.0}\tiny{(0.0)}  & 83.3\tiny{(4.8)} & 95.6\tiny{(4.5)} & 81.2\tiny{(22.9)} & 91.2\tiny{(3.2)} & /& \textbf{100.0}\tiny{(0.0)} & 2e6\\
    \tiny{2c vs 64zg} & Hard & \textbf{100.0}\tiny{(3.1)} & \textbf{100.0}\tiny{(0.0)}  &  \textbf{100.0}\tiny{(3.1)} & \textbf{100.0}\tiny{(2.7)} & 90.0\tiny{(4.8)} & 90.3\tiny{(4.0)} & / & \textbf{100.0}\tiny{(3.1)} & 5e6\\
    \tiny{3s5z} & Hard & 93.8\tiny{(2.6)} & \textbf{100.0}\tiny{(0.0)} & 74.0\tiny{(6.4)} & 72.5\tiny{(26.5)} & 90.0\tiny{(3.5)} & 84.3\tiny{(5.4)} & / & \textbf{100.0}\tiny{(3.1)} & 3e6\\
    \tiny{5m vs 6m} & Hard & 92.7\tiny{(1.5)} & 89.6\tiny{(1.5)} & 81.3\tiny{(5.1)} & 88.2\tiny{(6.2)} & 73.8\tiny{(4.4)} & 75.8\tiny{(3.7)} & 90.6\tiny{(6.1)}& \textbf{93.8}\tiny{(4.4)} & 1e7\\
    \tiny{8m vs 9m} & Hard & \textbf{100.0}\tiny{(1.5)} & \textbf{100.0}\tiny{(0.0)} &  96.9\tiny{(0.0)} & 93.8\tiny{(3.5)} & 86.2\tiny{(4.4)} & 92.6\tiny{(4.0)} & /& \textbf{100.0}\tiny{(3.1)} & 5e6\\
    \tiny{10m vs 11m} & Hard & \textbf{100.0}\tiny{(0.0)} & \textbf{100.0}\tiny{(1.4)} & \textbf{100.0}\tiny{(3.1)} & 96.3\tiny{(5.8)} & 77.5\tiny{(9.7)} & 95.8\tiny{(6.1)} & / & \textbf{100.0}\tiny{(0.0)} & 5e6\\
    \tiny{25m} & Hard & \textbf{100.0}\tiny{(0.0)} & \textbf{100.0}\tiny{(0.0)} &  0.0\tiny{(0.1)} & \textbf{100.0}\tiny{(2.7)} & 0.6\tiny{(0.8)} & 90.2\tiny{(9.8)} & 2.8\tiny{(3.1)}& \textbf{100.0}\tiny{(0.0)} & 2e6\\
    \tiny{27m vs 30m} & Hard+ & 96.9\tiny{(0.0)} & 96.9\tiny{(3.1)} & 80.2\tiny{(4.8)} & 93.1\tiny{(3.2)} & 0.0\tiny{(0.0)} & 39.2\tiny{(8.8)} & /& \textbf{100.0}\tiny{(0.0)} & 1e7\\
    \tiny{MMM2} & Hard+ & 92.7\tiny{(1.5)} & \textbf{100.0}\tiny{(3.1)} & 96.9\tiny{(0.0)} & 81.8\tiny{(10.1)} & 0.3\tiny{(0.4)} & 88.3\tiny{(2.4)} & / & \textbf{100.0}\tiny{(0.0)} & 1e7\\
    \tiny{6h vs 8z} & Hard+ & 83.3\tiny{(2.9)} & 96.9\tiny{(3.1)}  & 93.8\tiny{(4.4)} & 88.4\tiny{(5.7)} & 0.0\tiny{(0.0)} & 9.7\tiny{(3.1)} & /& \textbf{100.0}\tiny{(0.0)} & 1e7\\
    \tiny{3s5z vs 3s6z} & Hard+ & 78.1\tiny{(7.8)} & 87.5\tiny{(3.1)} & 79.2\tiny{(9.0)} & 84.3\tiny{(19.4)} & 82.8\tiny{(21.2)} & 68.8\tiny{(21.2)} & /& \textbf{100.0}\tiny{(3.1)} & 2e7\\
    \bottomrule
    \toprule
    Task & Difficulty & CF-dyn(0.4) & CF(0.4) & QGNN & SMS & TarMAC & NDQ & MAGIC & QMIX  & Steps\\
    \midrule
    \tiny{1o2r vs 4r} & Hard+ &  88.5\tiny{(1.5)} & \textbf{96.9}\tiny{(1.5)} & 93.8\tiny{(2.6)} & 76.4 & 39.1 & 77.1 & 22.3 & 51.1  & 2e7  \\
    \tiny{5z vs 1ul} & Hard+ & 89.6\tiny{(1.5)} & \textbf{100.0}\tiny{(1.4)} & 92.2\tiny{(1.6)} & 59.9 & 44.2 & 48.9 & 0.0 & 82.6 & 1e7 \\
    \tiny{1o10b vs 1r} & Hard+ & 84.4\tiny{(3.1)} & 96.9\tiny{(3.1)} & \textbf{98.0}\tiny{(2.9)} & 86.0 & 40.1 & 78.1 & 5.8 & 51.4 & 2e7 \\
    \bottomrule
    \toprule
    Task & Metric & CF-dyn(0.4) & CF(0.4) & MAGIC & HetNet & CommNet & I3CNet & TarMAC & GA-Comm & Steps\\
    \midrule
     & Success Rate & \textbf{100.0}\tiny{(0.0)} & \textbf{100.0}\tiny{(0.0)} & 98.2\tiny{(1.0)} & / & 59.2\tiny{(13.7)} & 70.0\tiny{(9.8)} & 73.5\tiny{(8.3)} & 88.8\tiny{(3.9)} & - \\
    
    \multirow{-2}{*}{GRF}& Steps Taken & 27.3\tiny{(0.3)} & \textbf{25.4}\tiny{(0.4)} & 34.3\tiny{(1.3)} & / & 39.3\tiny{(2.4)} & 40.4\tiny{(1.2)} & 41.5\tiny{(2.8)} & 39.1\tiny{(3.1)} & - \\ \midrule
     & Avg. Cumulative $\mathcal{R}$ & \textbf{-0.117}\tiny{(0.006)} & -0.121\tiny{(0.008)}  & -0.386\tiny{(0.024)}  & -0.232\tiny{(0.010)} & -0.336\tiny{(0.012)} & -0.342\tiny{(0.015)} & -0.563\tiny{(0.030)} & / & - \\
    \multirow{-2}{*}{PP}& Steps Taken & \textbf{4.82}\tiny{(0.21)} & 4.99\tiny{(0.31)} & 10.6\tiny{(0.50)} & 8.30\tiny{(0.25)} & 8.97\tiny{(0.25)} & 9.69\tiny{(0.26)} & 18.4\tiny{(0.46)} & / & - \\ \midrule
    & Avg. Cumulative $\mathcal{R}$ & -0.389\tiny{(0.007)} & \textbf{-0.197}\tiny{(0.019)} & -0.394\tiny{(0.017)} & -0.364\tiny{(0.017)} & -0.394\tiny{(0.019)} & -0.411\tiny{(0.019)} & -0.548\tiny{(0.031)} & / & - \\
    \multirow{-2}{*}{PCP}& Steps Taken & 15.7\tiny{(0.17)} & \textbf{7.61}\tiny{(0.66)} & 10.8\tiny{(0.45)} & 9.98\tiny{(0.36)} & 11.3\tiny{(0.34)} & 11.5\tiny{(0.37)} & 17.0\tiny{(0.80)} & / & - \\
    \bottomrule
    \end{tabular}
    }
\label{tab:smac}
\end{table*}

\subsection{Baselines}

We compare CommFormer against strong non-communication baselines and popular communication methods to highlight its effectiveness. All baselines follow the CTDE paradigm to ensure a fair comparison.

The strong MARL baselines, which primarily focus on algorithm framework design, include:
(1) \textbf{MAPPO} \citep{mappo}, which applies PPO in MARL using a shared set of parameters for all agents, without communication.
(2) \textbf{HAPPO} \citep{kuba2021trust}, which implements multi-agent trust-region learning with a sequential update scheme and monotonic improvement guarantees. 
(3) \textbf{QMIX} \citep{rashid2020monotonic}, which incorporates a centralized value function to facilitate decentralized decision-making and address credit assignment issues. 
(4) \textbf{UPDeT} \citep{hu2021updet}, which decouples agent observations into sequences of observation entities and uses a Transformer to map these sequences to actions. 
(5) \textbf{MAT} \citep{wen2022multi}, which treats cooperative MARL as sequence modeling, using a fixed encoder and fully decentralized actors for each agent.

The popular communication methods, which focus on learning communication strategies, include:
(1) \textbf{SMS} \citep{xue2022efficient}, which calculates the Shapley Message Value to evaluate the importance of each message and learn what to communicate.
(2) \textbf{TarMAC} \citep{das2019tarmac}, which uses an attention mechanism to integrate messages based on their importance.
(3) \textbf{NDQ} \citep{wang2019learning}, which learns nearly decomposable Q-functions with minimal communication, allowing agents to act independently and occasionally exchange messages for coordination. 
(4) \textbf{MAGIC} \citep{niu2021multi}, which employs hard attention to construct a dynamic communication graph and uses graph attention networks to process messages. 
(5) \textbf{QGNN} \citep{kortvelesy2022qgnn}, which introduces a value factorization method using a graph neural network with multi-layer message passing to learn what to communicate.
(6) \textbf{CommNet} \citep{sukhbaatar2016learning}, which uses continuous communication for fully cooperative tasks, learning what to communicate for each agent.
(7) \textbf{I3CNet} \citep{singh2018learning}, which controls continuous communication with a gating mechanism to learn when to communicate, constructing a dynamic communication graph.
(8) \textbf{GA-Comm} \citep{liu2020multi}, which models agent relationships using a complete graph and learns both whether and how agents should interact.
(9) \textbf{HetNet} \citep{seraj2022learning}, which learns efficient and diverse communication models based on heterogeneous graph attention networks, focusing on what to communicate.

We also include the \textbf{fully connected CommFormer} (FC) configuration, where each agent can communicate with all others, implying that the sparsity parameter $\mathcal{S}$ is set to 1.
This serves as an upper bound for our method, demonstrating strong performance in cooperative MARL tasks.
In our experiments, the implementations of baseline methods remain consistent with their official repositories, with all hyperparameters left unchanged at their original best-performing configurations. Details of these hyperparameters are provided in the Appendix.

Note that our approach is divided into two versions: the first, \textbf{CommFormer}, maintains a static communication graph during inference, where the communication structure remains fixed once established. The second version, \textbf{CommFormer-dyn}, dynamically adjusts the communication graph based on the outputs of our dynamic gating network during inference.

\subsection{Main Results}

According to the results presented in Table \ref{tab:smac} and Figure \ref{fig:variance}, our CommFormer with a sparsity parameter $\mathcal{S}=0.4$ significantly outperforms the state-of-the-art baselines. 
It consistently finds the optimal communication architecture across diverse cooperative scenarios, regardless of changes in the number of agents.
Take the task \textit{3s5z} as an example, where the algorithm needs to control different types of agents: stalkers and zealots.
This requires careful design of the communication architecture based on the capabilities of different units. 
Otherwise, it can even have a detrimental impact on performance, as indicated by the substantial variance displayed in Figure \ref{fig:variance}.
The outcome of \textit{3s5z} presented in Table \ref{tab:smac} consistently highlights CommFormer's ability to attain optimal performance with different random seeds, which underscores the robustness and efficiency of our proposed method.
Furthermore, in comparison to the FC method, CommFormer nearly matches its performance while retaining only 40\% of the edges. 
This indicates that with an appropriate communication architecture, many communication channels can be eliminated, thereby significantly reducing the hardware communication equipment requirements and expanding its applicability.
Finally, it's worth noting that all the results presented in Table \ref{tab:smac} are based on the same number of training steps, demonstrating the robustness and effectiveness of our bi-level optimization approach, which consistently converges to the optimal solution while maintaining sample efficiency. 

CommFormer-dyn, which infers with a dynamic communication graph, achieves performance comparable to the static communication graph version in most scenarios. 
However, in certain complex tasks, such as \textit{6h\_vs\_8z}, \textit{3s5z\_vs\_3s6z} in SMAC settings, and PCP tasks, a performance gap remains, indicating challenges in optimally deciding when to communicate without sacrificing performance. 
Nevertheless, when compared to other dynamic communication methods, such as I3CNet and MAGIC, our approach significantly outperforms these baselines, demonstrating the efficiency of our framework. 
This indicates that even using only current observations is sufficient to effectively determine communication timing in most tasks from a graph modeling perspective.
The hyper-parameters used in the study, along with additional detailed results, are provided in the Appendix.

\subsection{Ablations}

We conduct several ablation studies, primarily focusing on the SMAC environments, to examine specific aspects of our CommFormer.
The parameter $\mathcal{S}$, which determines the sparsity of the communication graph, impacts the number of connected edges. 
Lower values of $\mathcal{S}$ imply reduced costs associated with communication but may also lead to performance degradation.
Additionally, we conduct ablation studies to investigate the essence of architecture searching, where we generate various pre-defined architectures using different random seeds, simulating manually pre-defined settings. 
Furthermore, we visualize the edge-learning process to illustrate how the communication graph evolves during training.
Lastly, we analyze the dynamic gating mechanism and its activation process, a core component of CommFormer-dyn.
These ablation studies serve to enhance our understanding of the underlying motivations of our method and provide valuable insights into its key components and their impact on performance.
Note that, aside from the dynamic gating and sparsity ablation studies, all other experiments are conducted solely on CommFormer with a static communication graph during inference.

\begin{figure*}
    \centering
    \subfloat[]{\includegraphics[width=1.7in]{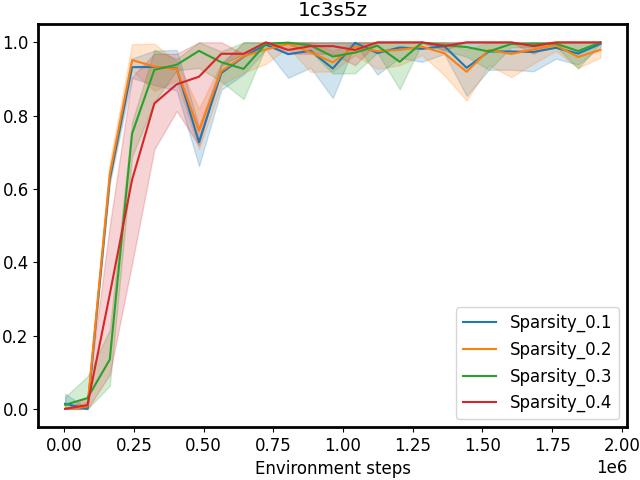}}
    \hfil
    \subfloat[]{\includegraphics[width=1.7in]{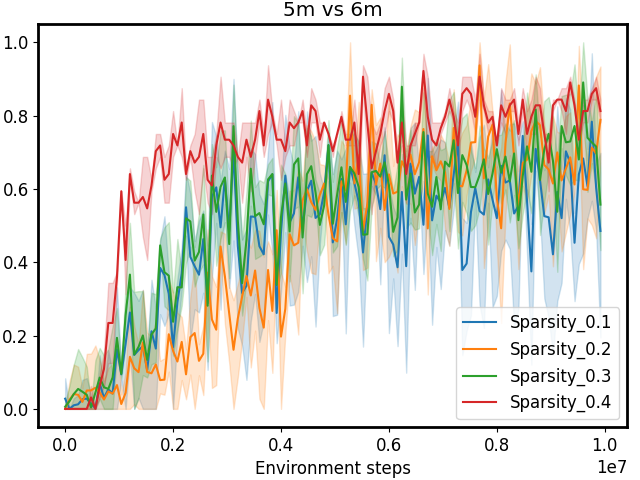} }
    \hfil
    \subfloat[]{\includegraphics[width=1.7in]{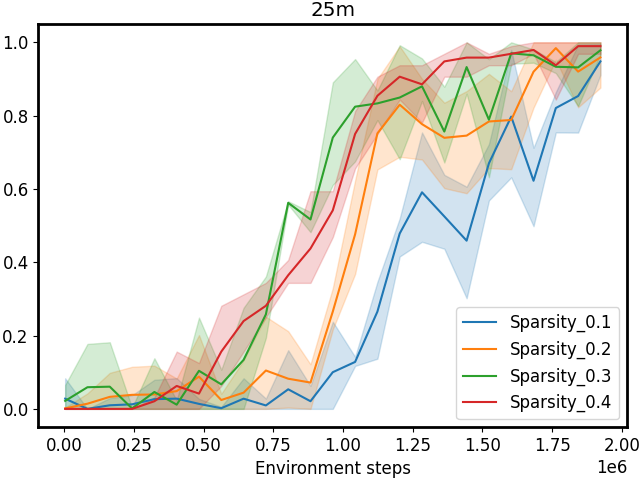} }
    \hfil \\
    \subfloat[]{\includegraphics[width=1.7in]{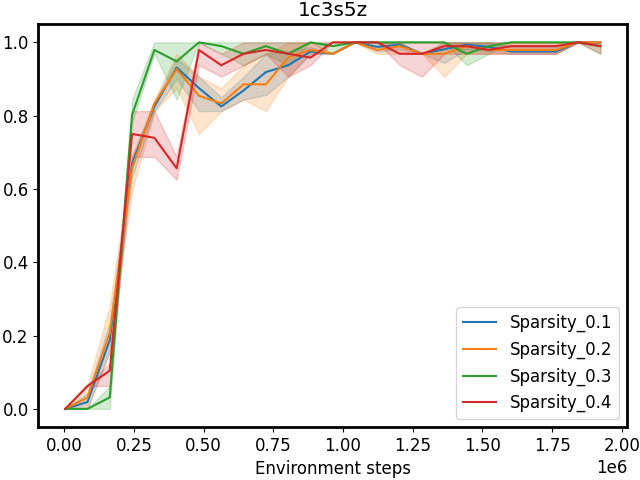}}
    \hfil
    \subfloat[]{\includegraphics[width=1.7in]{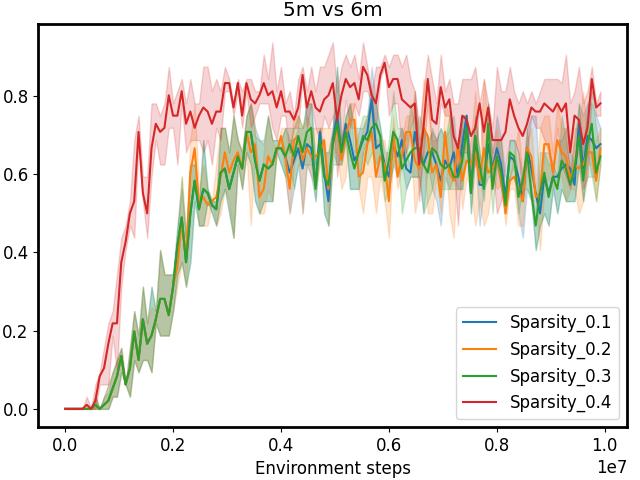} }
    \hfil
    \subfloat[]{\includegraphics[width=1.7in]{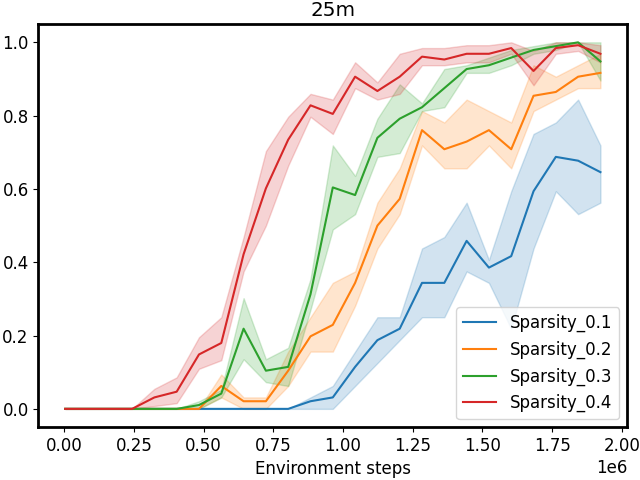} }
    \caption{
    Performance comparison on SMAC tasks \textit{1c3s5z}, \textit{5m\_vs\_6m}, and \textit{25m} with different sparsity $\mathcal{S}$.
    The first row shows results for CommFormer, while the second row presents results for CommFormer-dyn, which includes an additional dynamic gating mechanism.
    As the value of sparsity $\mathcal{S}$ increases, both CommFormer and CommFormer-dyn show improved performance across different environments, with this effect being particularly pronounced in scenarios involving a large number of agents. 
    }
    \label{fig:absparsity}
\end{figure*}

\begin{figure*}
    \centering
    \begin{subfigure}{0.32\linewidth}
        \centering
        \includegraphics[width=1.7in]{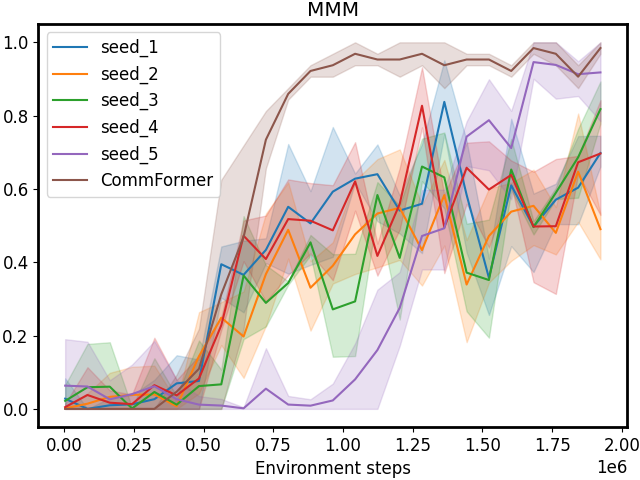} 
    \end{subfigure}%
    \begin{subfigure}{0.32\linewidth}
        \centering
        \includegraphics[width=1.7in]{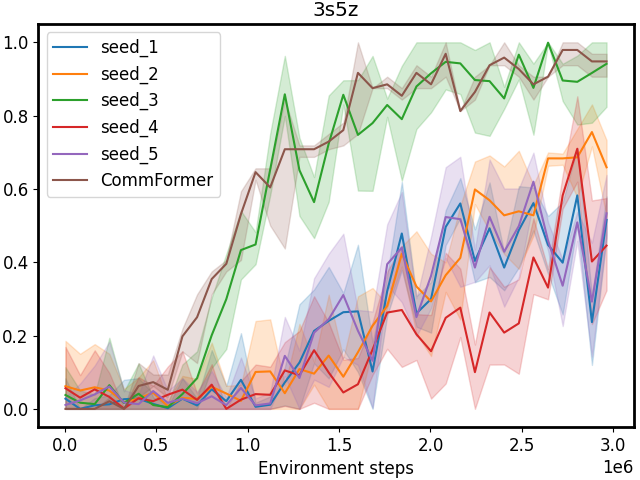} 
    \end{subfigure}%
    \begin{subfigure}{0.32\linewidth}
        \centering
        \includegraphics[width=1.7in]{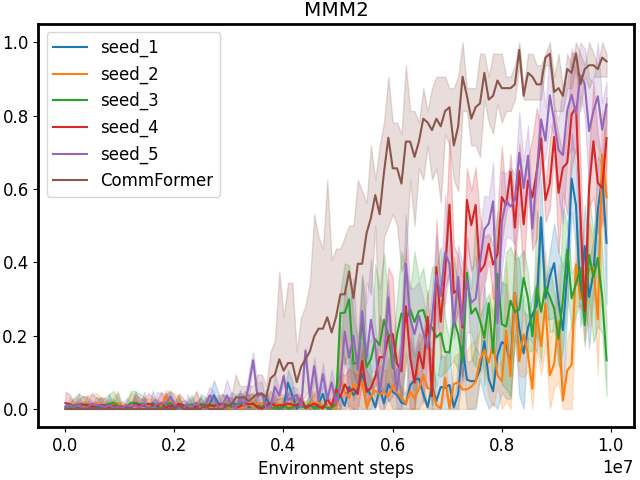} 
    \end{subfigure}%
    \caption{
    Performance comparison on SMAC tasks with different manually pre-defined communication architectures.
    CommFormer consistently achieves optimal performance, which underscores its capability to autonomously search for the optimal communication architecture, highlighting its adaptability across various scenarios and tasks.
    }
    \label{fig:absearch}
\end{figure*}

\textbf{Sparsity.}
The parameter $\mathcal{S}$ introduced in our bi-level optimization controls the number of connected edges, ensuring that it does not exceed $\mathcal{S} \times N^2$, as specified in Equation \ref{eq_sparsity}. 
To simplify this constraint, we ensure that the total number of edges $|\alpha|$ equals $\mathcal{S} \times N^2$, with each agent communicating with a fixed number of $\mathcal{S} \times N$ agents.
Smaller values of $\mathcal{S}$ reduce the cost associated with communication but may also result in performance degradation. 
To investigate the impact of varying $\mathcal{S}$, we conduct a series of experiments with both CommFormer and CommFormer-dyn, whose results are presented in Figure \ref{fig:absparsity}.
The observed trend is consistent across both methods:
for simpler tasks, such as \textit{1c3s5z}, achieving a 100\% win rate is possible even when each agent can only communicate with one other agent. 
Nevertheless, as task complexity and the number of participating agents increase, a larger value of sparsity $\mathcal{S}$ becomes necessary to attain superior performance.

\textbf{Architecture Searching.} 
In light of the constraints imposed by limited bandwidth and contention for medium access, designing the communication architecture for each agent becomes a critical task.
To investigate the impact of different communication graph configurations, we conduct experiments using various random seeds to generate diverse configurations, simulating the effects of individual configuration choices on the problem.
The results, as depicted in Figure \ref{fig:variance} and \ref{fig:absearch}, highlight that manually pre-defining the communication architecture often leads to significant performance variance, demanding expert knowledge for achieving better results.
In contrast, our proposed method leverages the continuous relaxation of the graph representation. This innovative approach allows for the simultaneous optimization of both the communication graph and architectural parameters in an end-to-end fashion, all while maintaining sample efficiency. 
This underscores the essentiality and effectiveness of our approach in tackling the challenges of multi-agent communication in constrained environments.

\begin{figure}[!tbp]
    \centering
    \includegraphics[width=1.0\linewidth]{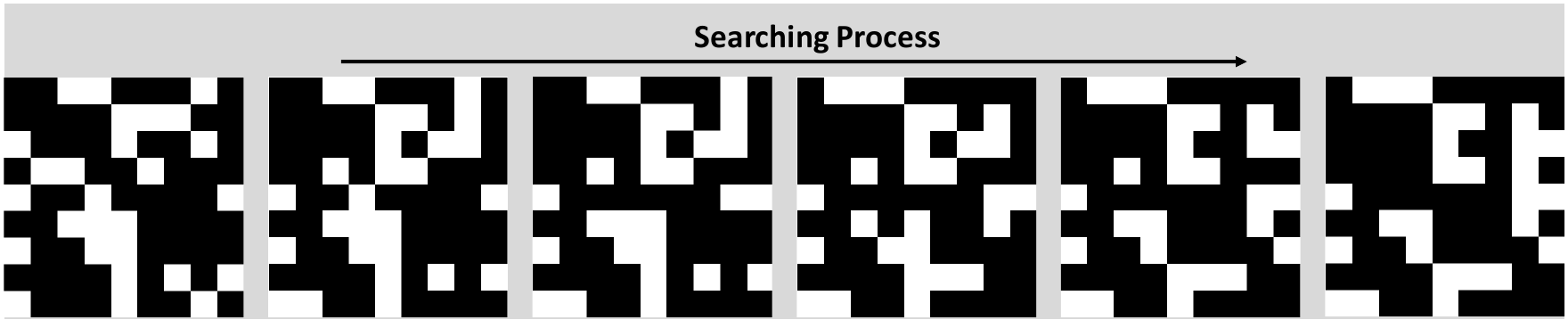}
    \caption{The searching process of CommFormer in the SMAC task \textit{1c3s5z}. In this representation, a white square corresponds to a value of 1, indicating the presence of an edge connection.  }
    \label{fig:search}
\end{figure}

\textbf{Visualization of Edge Learning.} 
We present additional visual results (Figure \ref{fig:search})  that showcase the final communication architecture obtained through our search process. 
These visualizations offer a more intuitive understanding of the architecture's evolution during training. 
In the initial stages, the communication graph undergoes substantial modifications, adapting to enhance overall model performance through significant changes in edge connectivity and agent interactions.
In the later stages of training, the architecture stabilizes, exhibiting only minimal adjustments—typically involving the addition or removal of one or two edges—as both the communication graph and model performance reach a steady state.

\textbf{Dynamic Gating Input.} 
In our framework, we develop a dynamic gating mechanism for each agent to determine whether to receive shared information from the sender, as specified by the communication graph, based on current observations. 
In previous studies, the input to the gating mechanism often includes historical information \citep{singh2018learning, niu2021multi}, typically encoded by recurrent networks, such as LSTMs or RNNs, to decide whether the current decision requires shared information.
We similarly incorporate historical information into the input of gating network, which can be formulated as:
\begin{equation}
\label{eq:dyn_h2}
    h^i_t = \Psi^{\text{gate}}(o^i_t, c^{i}_t | \kappa_i) ~~ \text{for} ~~ i=1,\dots, n,
\end{equation}
where $c^{i}_{t-1}$ represents the temporal hidden state from the previous time step $t-1$ for agent $i$, captured via an RNN. 
We conduct several ablation studies to assess the effectiveness of  Equation \ref{eq:dyn_h2}, as presented in Table \ref{tab:ab_input}.
As task difficulty increases, Equation \ref{eq:dyn_h2} performs worse than Equation \ref{eq:dyn_h}, suggesting that in complex tasks, basing communication decisions solely on the current observation is sufficient within our framework.
Additional historical information may even hinder communication decisions by introducing potentially overly optimistic past data, which can include similar states that do not require communication, thus misguiding the current judgment.
This approach also reduces the burden on each agent by eliminating the need to store historical information.

\begin{table}[t!]
\setlength{\tabcolsep}{4pt}
\centering
\caption{Ablation study on dynamic gating using different inputs for the gating network, based on Equations \ref{eq:dyn_h} and \ref{eq:dyn_h2}. The performance is evaluated using different metrics and standard deviations across the selected benchmark.}
\label{tab:ab_input}
\small
\centering
\scalebox{1.0}{
\begin{tabular}{C{1.4cm}C{1.4cm}|C{2.0cm}C{2.0cm}}
\toprule[2pt]
Task & Difficulty & Equation \ref{eq:dyn_h} & Equation \ref{eq:dyn_h2} \\
\midrule 
MMM & Easy & 100.0\tiny{(0.0)} & 100.0\tiny{(1.5)} \\
5m vs 6m & Hard & 92.7\tiny{(1.5)} & 66.7\tiny{(1.5)}\\
25m & Hard & 100.0\tiny{(0.0)} & 75.0\tiny{(5.1)} \\
MMM2 & Hard+ & 92.7\tiny{(1.5)} & 43.8\tiny{7.7} \\
6h vs 8z & Hard+ & 83.3\tiny{(2.9)} & 46.9\tiny{6.8} \\
\bottomrule[2pt]
\end{tabular}
}
\end{table}

\begin{figure}[!tbp]
    \centering
    \includegraphics[width=1.0\linewidth]{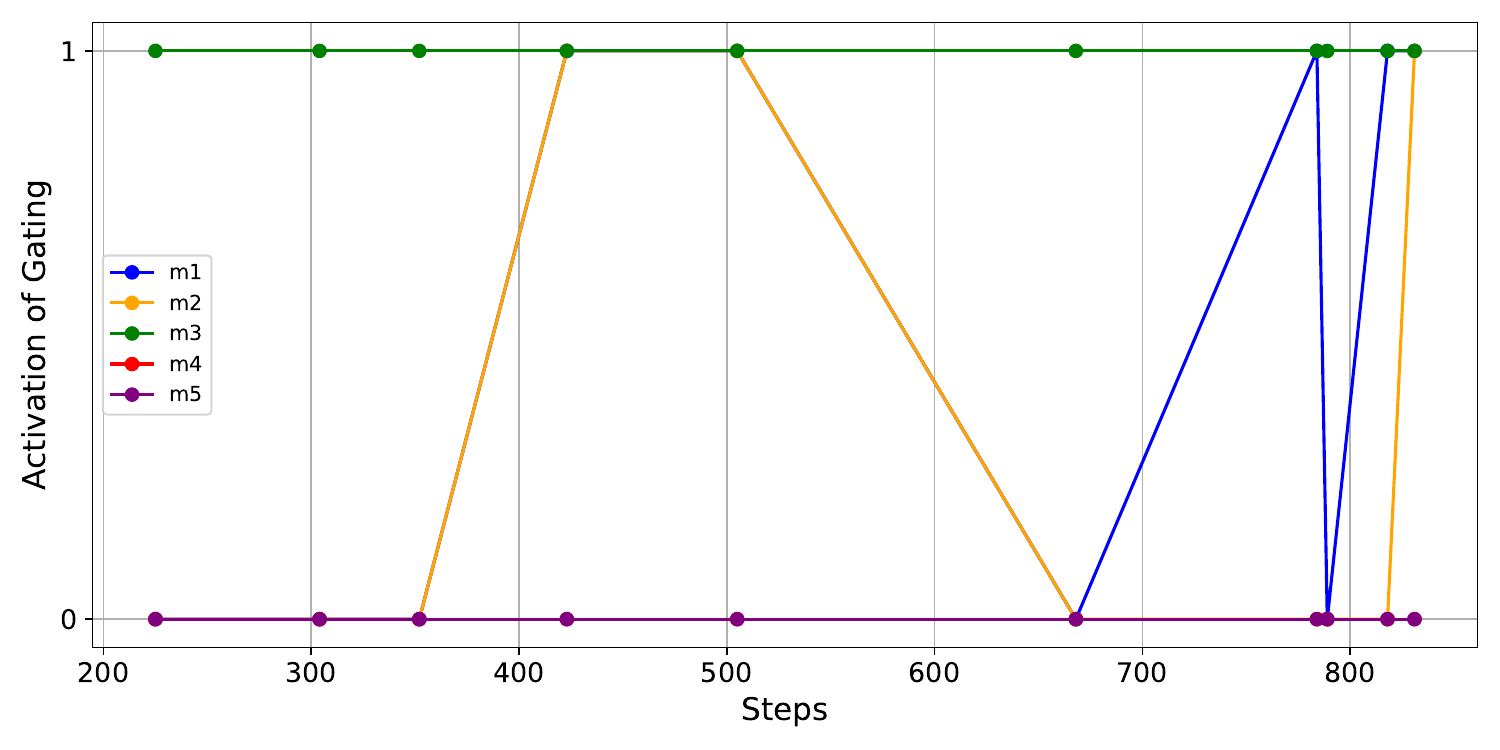}
    \caption{Activation process of the gating network in the \textit{5m\_vs\_6m} task. Note that agents $m4$ and $m5$ remain inactive, not requiring information from others, while $m3$ remains active and consistently relies on shared information for decision-making. Agents $m1$ and $m2$ dynamically decide to accept shared information based on the situational context.}
    \label{fig:inf_of_gating}
\end{figure}

\textbf{Dynamic Gating Activation.} 
Since in most cases only key decisions require information sharing among agents, we incorporate a temporal gating mechanism to identify critical moments for communication, significantly reducing resource consumption. 
To evaluate its effectiveness, we use the \textit{5m\_vs\_6m} task as an example, illustrating the activation process among the five Marines.
As shown in Figure \ref{fig:inf_of_gating}, agents refrain from communicating most of the time, yet still achieve superior performance ($92.7$) compared to a static communication graph ($89.6$), with considerably lower communication resource usage. 
Additionally, our method significantly outperforms the no-communication scenario ($81.3$), demonstrating its ability to effectively identify key moments for communication and validating its overall effectiveness.
\section{Conclusion}

In this paper, we introduce a novel approach called CommFormer, which addresses the challenge of learning multi-agent communication from a graph modeling perspective.
Our approach treats the communication architecture among agents as a learnable graph and formulates this problem as the task of determining the communication graph while enabling the architecture parameters to update normally, thus necessitating a bi-level optimization process.
By leveraging the continuous relaxation of graph representation and incorporating attention mechanisms within the graph modeling framework, CommFormer enables the concurrent optimization of the communication graph and architectural parameters in an end-to-end manner.
Additionally, we introduce a temporal gating mechanism for each agent, enabling dynamic decisions on whether to receive shared information at a given time, based on current observations, thus improving decision-making efficiency.
Extensive experiments conducted on a variety of cooperative tasks illustrate the significant performance advantage of our approach compared to other state-of-the-art baseline methods. 
In fact, CommFormer approaches the upper bound in scenarios where unrestricted information sharing among all agents is permitted, while CommFormer-dyn is able to maintain performance levels comparable to CommFormer in most scenarios.
We believe that our work opens up new possibilities for the application of communication learning in the field of MARL, where effective communication plays a pivotal role in addressing various challenges.

\bibliography{main}
\bibliographystyle{plainnat}

\clearpage
\onecolumn
\appendices

\section{Hyper-parameter Settings}
\label{sec:hyper}
During our experiments, we maintain consistency in the implementations of baseline methods by using their official repositories, and we keep all hyperparameters unchanged from their original best-performing configurations.
Specific hyperparameters used for different algorithms and tasks can be found in Tables \ref{table:common-smac} to \ref{tab:specific_SMAC}. 
To ensure a fair comparison and validate that our approach achieves optimal performance without compromising sample efficiency, we adopted the same hyperparameter settings as MAT \citep{wen2022multi}.

\section{Details of Experimental Results}
\begin{figure*}[!htbp]
 	\centering
 	\vspace{-5pt}
 	\begin{subfigure}{0.32\linewidth}
 		\centering
 		\includegraphics[width=1.7in]{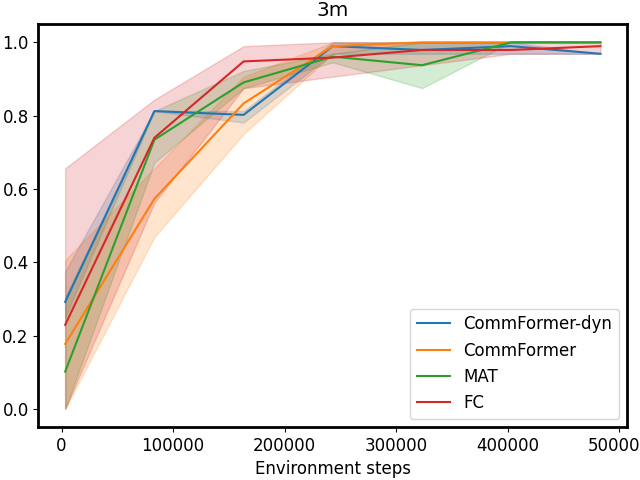} 
 	\end{subfigure}%
 	\begin{subfigure}{0.32\linewidth}
 		\centering
 		\includegraphics[width=1.7in]{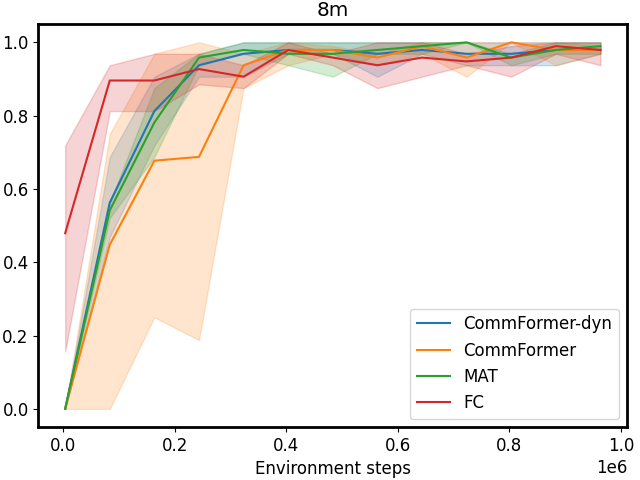} 
 	\end{subfigure}%
 	\begin{subfigure}{0.32\linewidth}
 		\centering
 		\includegraphics[width=1.7in]{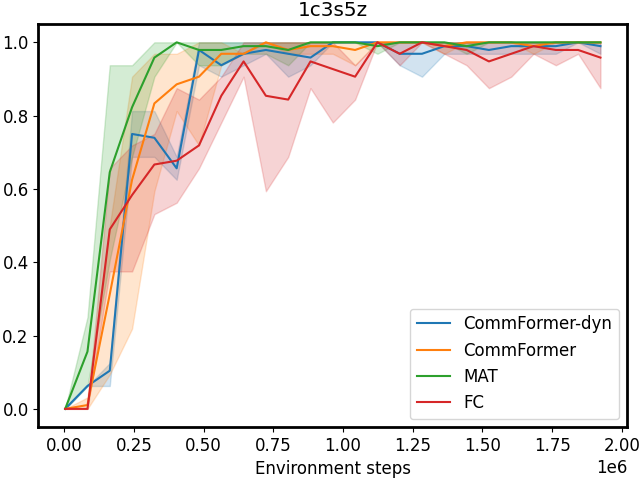} 
 	\end{subfigure}
 	\begin{subfigure}{0.32\linewidth}
 		\centering
 		\includegraphics[width=1.7in]{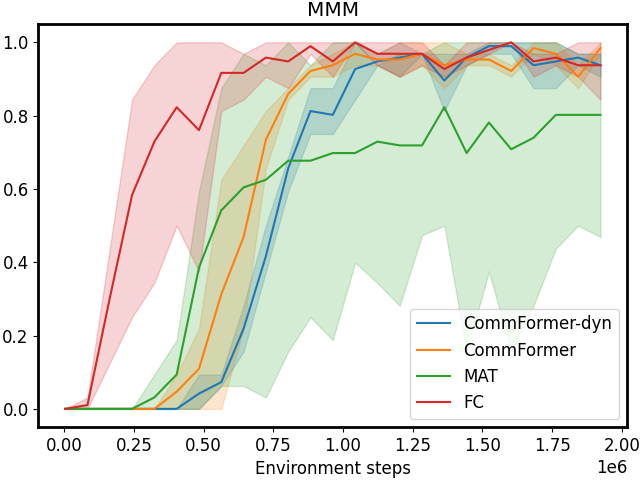}
 	\end{subfigure}%
 	\begin{subfigure}{0.32\linewidth}
 		\centering
 		\includegraphics[width=1.7in]{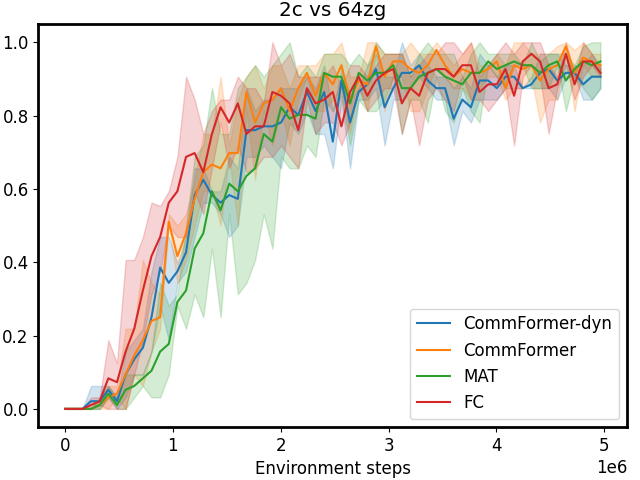}
 	\end{subfigure}%
 	\begin{subfigure}{0.32\linewidth}
 		\centering
 		\includegraphics[width=1.7in]{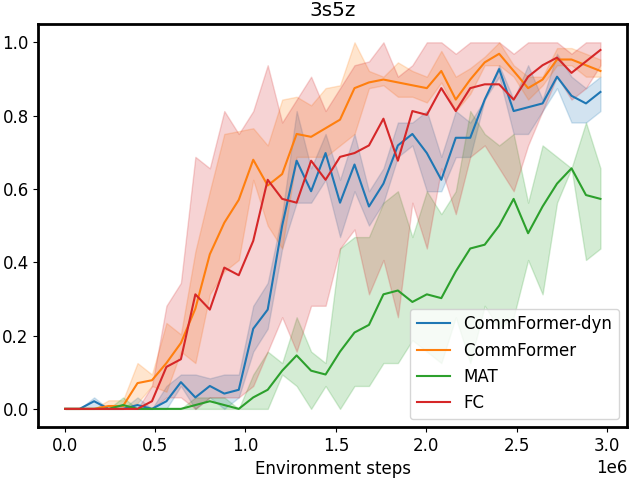}
 	\end{subfigure}
 	\begin{subfigure}{0.32\linewidth}
 		\centering
 		\includegraphics[width=1.7in]{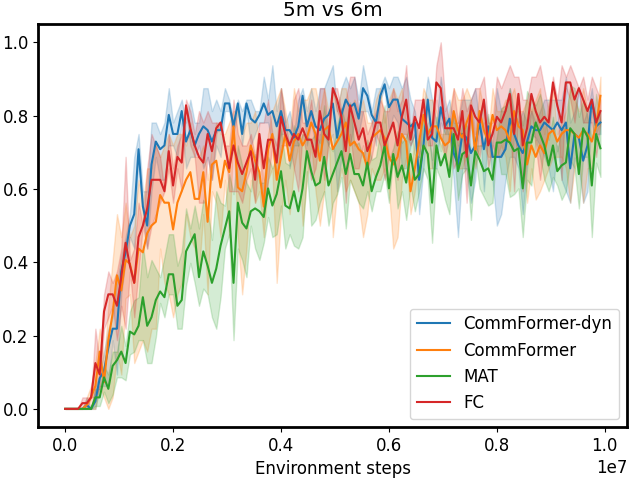}
 	\end{subfigure}%
 	\begin{subfigure}{0.32\linewidth}
 		\centering
 		\includegraphics[width=1.7in]{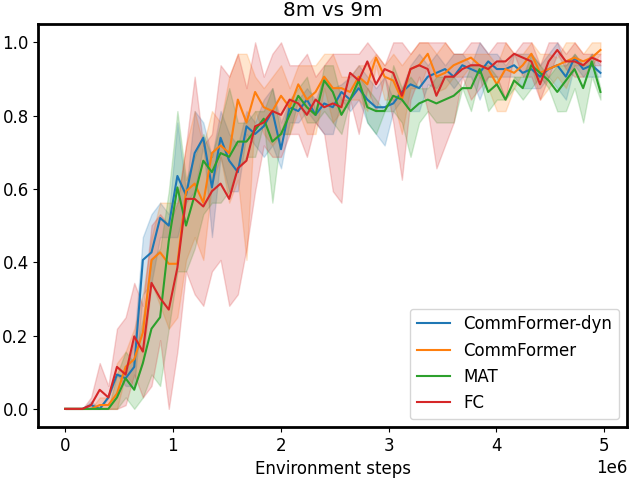}
 	\end{subfigure}%
 	\begin{subfigure}{0.32\linewidth}
 		\centering
 		\includegraphics[width=1.7in]{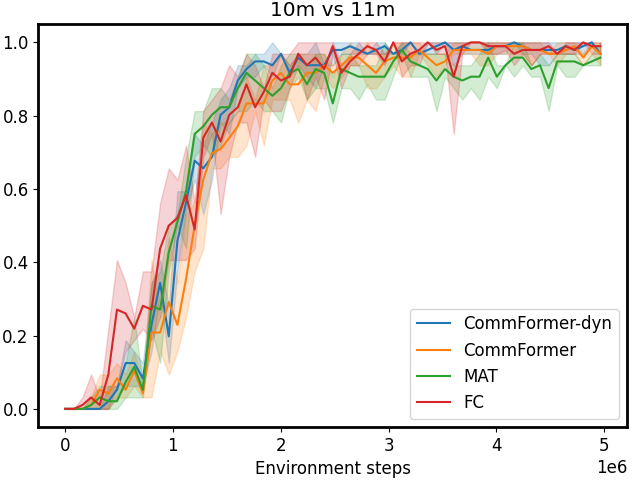}
 	\end{subfigure}
 	\begin{subfigure}{0.32\linewidth}
 		\centering
 		\includegraphics[width=1.7in]{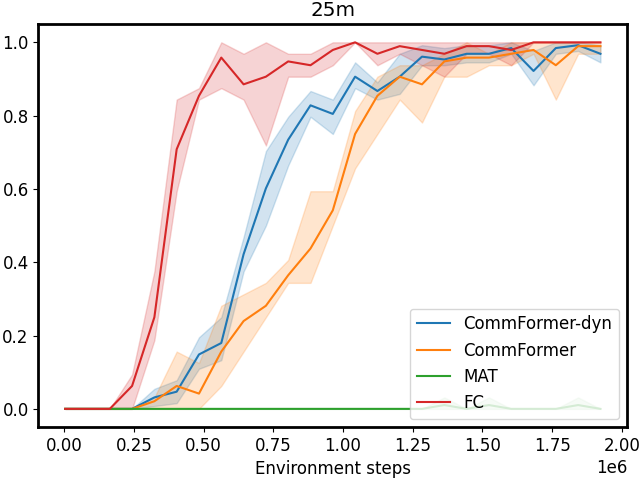} 
 	\end{subfigure}%
 	\begin{subfigure}{0.32\linewidth}
 		\centering
 		\includegraphics[width=1.7in]{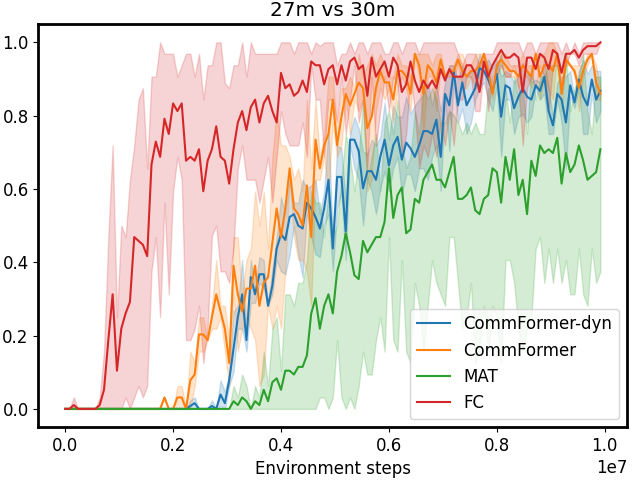}
 	\end{subfigure}%
 	\begin{subfigure}{0.32\linewidth}
 		\centering
 		\includegraphics[width=1.7in]{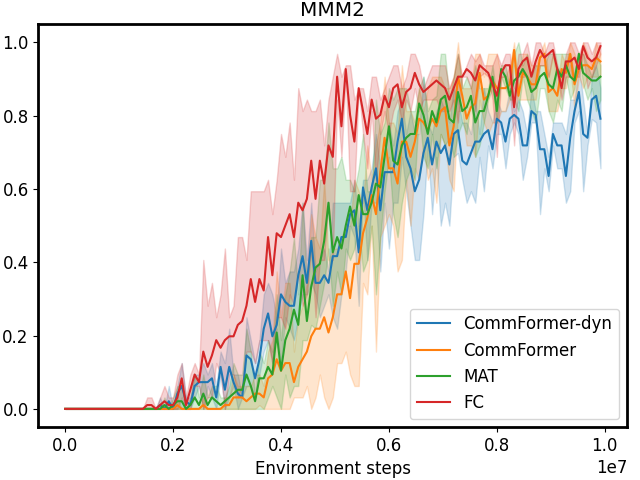} 
 	\end{subfigure}
 	\begin{subfigure}{0.45\linewidth}
 		\centering
 		\includegraphics[width=1.7in]{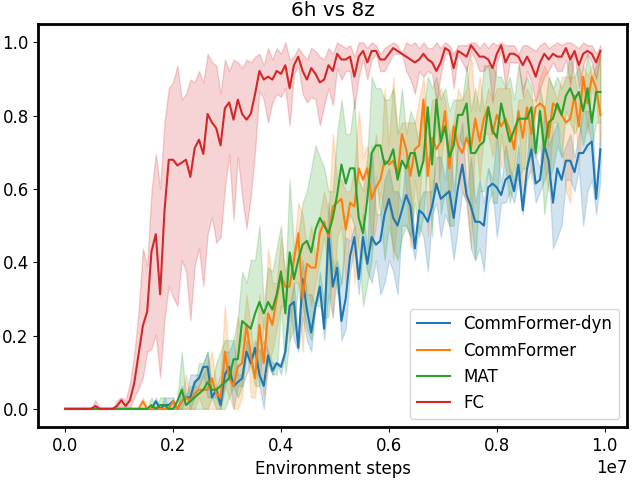}
 	\end{subfigure}%
 	\begin{subfigure}{0.45\linewidth}
 		\centering
 		\includegraphics[width=1.7in]{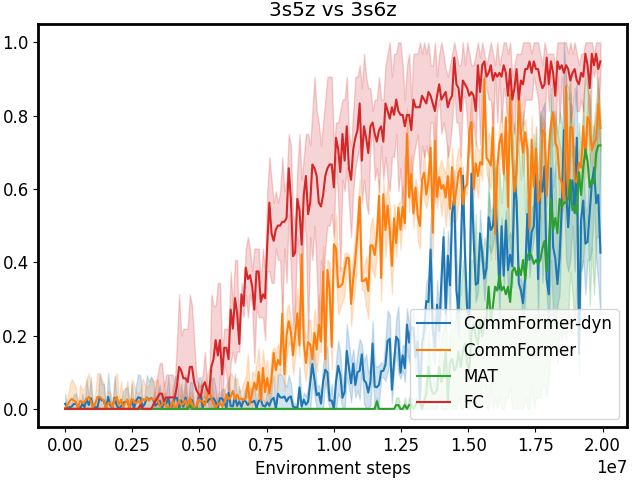} 
 	\end{subfigure}%
    \vspace{-0pt}
    \caption{\normalsize Performance comparison on SMAC tasks. Both CommFormer and CommFormer-dyn consistently outperform strong baselines and achieve performance levels comparable to methods that permit unrestricted information sharing among all agents, demonstrating their effectiveness across varying agent configurations. } 
    \label{fig:detail}
\end{figure*}

We provide detailed training figures (Figure \ref{fig:detail}) for various methods to substantiate our claim that our approach facilitates simultaneous optimization of the communication graph and architectural parameters in an end-to-end manner, all while preserving sample efficiency.

\section{Application Consideration}

A possible application of this study is to create an efficient communication framework tailored for enclosed, finite environments, typical of logistics warehouses. In these settings, agent movement is limited to designated zones, and communication is facilitated through overhead wires, akin to a trolleybus system.

In contrast, open environments present unique challenges, primarily due to the potential vast distances between agents, which require wireless communication and may hinder effective communication. 
To address this, a straightforward approach could be to add bidirectional edges between agents when they come within close proximity, enabling communication between them \citep{seraj2022learning}. 
However, a more effective solution may involve a hybrid approach that considers the constraint on the available bandwidth: initially segmenting agents into groups based on proximity, followed by an internal search for an optimal communication graph within each group. 
If agent distances vary dynamically during testing, this process is repeated as necessary to adjust the communication graph in real time, ensuring continuous adaptability to changing environmental conditions.

\newpage
\begin{table*}[!htbp]
\caption{Common hyper-parameters used for our method in the experiments.}
 \renewcommand{\arraystretch}{1.2}
  \centering
    \begin{tabular}{cc|cc|cc}
    \toprule
    hyper-parameters & value & hyper-parameters & value & hyper-parameters & value\\
    \midrule
    critic lr & 5e-4 & actor lr & 5e-4 & use gae & True\\
    gain & 0.01 & optim eps & 1e-5 & batch size & 3200\\
    training threads & 16 & num mini-batch & 1 & rollout threads & 32\\
    entropy coef & 0.01 & max grad norm & 10 & episode length & 100\\
    optimizer & Adam & hidden layer dim & 64 & use huber loss & True\\
    \bottomrule
    \end{tabular}
    \label{table:common-smac}
\end{table*}

\begin{table*}[!htbp]
\caption{Specific hyper-parameters used for our method in the experiments.}
 \renewcommand{\arraystretch}{1.2}
  \centering
  \scalebox{0.9}{
    \begin{tabular}{cc|cc|cc}
    \toprule
    hyper-parameters in PP & value & hyper-parameters in PCP & value & hyper-parameters in GRF & value\\
    \midrule
    Number Agents & 3 & Number Predators & 2 & Number Agents & 3\\
    Number Enemies & 1 & Number Captures & 1 & eval episode length & 200 \\
    vision & 1 & Number Enemies & 1 & - & -\\
    eval episode length & 20 & vision & 1 & - & -\\
    - & - & eval episode length & 20 & - & -\\
    \bottomrule
    \end{tabular}}
    \label{table:common-other}
\end{table*}

\begin{table*}[!htbp]
\caption{Different hyper-parameters used for CommFormer in different tasks.}
 \renewcommand{\arraystretch}{1.2}
  \centering
  \label{table2}
  \resizebox{0.8\textwidth}{!}{
    \begin{tabular}{c|ccccccc}
    \toprule
    tasks & ppo epochs & ppo clip & num blocks & num heads & stacked frames & steps & $\gamma$ \\
    \midrule
    3m & 15 & 0.2 & 1 & 1 & 1 & 5e5 & 0.99 \\
    8m & 15 & 0.2 & 1 & 1 & 1 & 1e6 & 0.99 \\
    1c3s5z & 10 & 0.2 & 1 & 1 & 1 & 2e6 & 0.99 \\
    MMM & 15 & 0.2 & 1 & 1 & 1 & 2e6 & 0.99 \\
    2c vs 64zg & 10 & 0.05 & 1 & 1 & 1 & 5e6 & 0.99  \\
    3s vs 5z & 15 & 0.05 & 1 & 1 & 4 & 5e6 & 0.99  \\ 
    3s5z & 10 & 0.05 & 1 & 1 & 1 & 3e6 & 0.99  \\
    5m vs 6m & 10 & 0.05 & 1 & 1 & 1 & 1e7 & 0.99  \\
    8m vs 9m & 10 & 0.05 & 1 & 1 & 1 & 5e6 & 0.99  \\
    10m vs 11m & 10 & 0.05 & 1 & 1 & 1 &  5e6 & 0.99 \\
    25m & 15 & 0.05 & 1 & 1 & 1 & 2e6 & 0.99 \\
    27m vs 30m & 5 & 0.2 & 1 & 1 & 1 & 1e7 & 0.99 \\
    MMM2 & 10 & 0.05 & 1 & 1 & 1 & 1e7 & 0.99 \\
    6h vs 8z & 15 & 0.05 & 1 & 1 & 1 & 1e7 & 0.99 \\
    3s5z vs 3s6z & 5 & 0.05 & 1 & 1 & 1 & 2e7 & 0.99 \\
    1o10b vs 1r & 10 & 0.2 & 1 & 1 & 1 & 2e7 & 0.99 \\
    1o2r vs 4r & 5 & 0.05 & 1 & 1 & 1 & 1e7 & 0.99 \\
    5z vs 1ul & 10 & 0.05 & 1 & 1 & 1 & 1e7 & 0.99 \\
    PP & 10 & 0.05 & 1 & 1 & 1 & 1e7 & 0.99 \\
    PCP & 10 & 0.05 & 1 & 1 & 1 & 1e7 & 0.99 \\
    GRF & 10 & 0.05 & 1 & 1 & 1 & 1e7 & 0.99 \\
    \bottomrule
    \end{tabular}
    }
    \label{tab:specific_SMAC}
\end{table*}

\vfill

\end{document}